\documentclass{article}
\pdfoutput=1 

\usepackage{arxiv}
\usepackage[utf8]{inputenc} 
\usepackage[T1]{fontenc}
\usepackage{lmodern}
\usepackage{hyperref}       
\usepackage{bookmark}       
\usepackage{url}            
\usepackage{booktabs}       
\usepackage{amsfonts}       
\usepackage{amsmath}
\usepackage{siunitx}
\usepackage{nicefrac}       
\usepackage{microtype}      
\usepackage{cleveref}       
\usepackage{graphicx}
\usepackage{natbib}
\usepackage{doi}
\usepackage{xcolor}
\usepackage{tikz}
\usepackage{pgfplots}
\usepackage{subcaption}
\usepackage{tcolorbox}
\usepackage{makecell}
\usepackage{adjustbox}
\usepackage{booktabs,multirow,arydshln}
\usepackage{fontawesome5}
\usepackage{dsfont}                 
\usetikzlibrary{shapes.geometric}
\usetikzlibrary{decorations.pathreplacing}
\usetikzlibrary{patterns}
\usetikzlibrary{positioning}
\pgfplotsset{compat=1.18}
\usepgfplotslibrary{groupplots}

\newcommand{\GVN}{500}
\newcommand{\GVGf}{0.458}
\newcommand{\GVAlbAcc}{0.516}

\newcommand{\GVRelAcc}{0.5}

\newcommand{\GVRghAcc}{0.446}

\newcommand{\GVRatAcc}{0.369}

\newcommand{\GVNaming}{0.858}
\newcommand{\GVShuffle}{0.52}
\newcommand{\GVZero}{0.631}

\newcommand{\GVCov}{0.73}

\newcommand{\GVMareN}{230}
\newcommand{\GVHl}{0.015}
\newcommand{\GVHlN}{270}
\newcommand{\GVEpAcc}{0.659}

\newcommand{\GVArMae}{0.425}
\newcommand{\GVSpearman}{0.295}
\newcommand{\GVConstMae}{0.43}

\newcommand{\GVCrep}{0.004}
\newcommand{\GVDOne}{0.78}
\newcommand{\GVGfMare}{0.527}
\newcommand{\GVGfHighland}{0.399}

\newcommand{\BLGemmaFourN}{120}
\newcommand{\BLGemmaFourGf}{0.176}
\newcommand{\BLGemmaFourAlbAcc}{0.283}

\newcommand{\BLGemmaFourRatAcc}{0}

\newcommand{\BLGemmaFourNaming}{0.8}

\newcommand{\BLGemmaFourCov}{1}

\newcommand{\BLGemmaFourHl}{0.268}

\newcommand{\BLGemmaFourEpAcc}{0.674}

\newcommand{\BLGemmaFourBeats}{no}

\newcommand{\BLQwenThreeFiveGf}{0.116}
\newcommand{\BLQwenThreeFiveAlbAcc}{0.092}

\newcommand{\BLQwenThreeFiveRatAcc}{0.038}

\newcommand{\BLQwenThreeFiveNaming}{0.833}

\newcommand{\BLQwenThreeFiveCov}{0.969}

\newcommand{\BLQwenThreeFiveHl}{0.839}

\newcommand{\BLQwenThreeFiveEpAcc}{0.513}

\newcommand{\BLQwenThreeFiveBeats}{no}

\newcommand{\BLDeepseekvlTwoGf}{0.165}
\newcommand{\BLDeepseekvlTwoAlbAcc}{0.258}

\newcommand{\BLDeepseekvlTwoRatAcc}{0.29}

\newcommand{\BLDeepseekvlTwoNaming}{0.642}

\newcommand{\BLDeepseekvlTwoCov}{0.125}

\newcommand{\BLDeepseekvlTwoHl}{0.161}

\newcommand{\BLDeepseekvlTwoEpAcc}{0.344}

\newcommand{\BLDeepseekvlTwoBeats}{no}

\newcommand{\BLGptFiveSixterraGf}{0.222}
\newcommand{\BLGptFiveSixterraAlbAcc}{0.533}

\newcommand{\BLGptFiveSixterraRatAcc}{0.042}

\newcommand{\BLGptFiveSixterraNaming}{0.783}

\newcommand{\BLGptFiveSixterraCov}{0.984}

\newcommand{\BLGptFiveSixterraHl}{0.679}

\newcommand{\BLGptFiveSixterraEpAcc}{0.717}

\newcommand{\BLGptFiveSixterraBeats}{no}

\newcommand{\BLClaudeopusFiveGf}{0.302}
\newcommand{\BLClaudeopusFiveAlbAcc}{0.548}

\newcommand{\BLClaudeopusFiveRatAcc}{0.195}

\newcommand{\BLClaudeopusFiveNaming}{0.85}

\newcommand{\BLClaudeopusFiveCov}{0.797}

\newcommand{\BLClaudeopusFiveHl}{0.107}

\newcommand{\BLClaudeopusFiveEpAcc}{0.304}

\newcommand{\BLClaudeopusFiveBeats}{no}

\newcommand{\ABIncGf}{0.494}

\newcommand{\PAKeeperRho}{0.387}
\newcommand{\PAKeeperSlope}{0.044}
\newcommand{\PAKeeperCov}{0.645}

\newcommand{\PACraterMae}{0.473}
\newcommand{\PACraterRho}{-0.21}

\newcommand{\PACraterCov}{0.731}

\newcommand{\PAConstMae}{0.407}

\newcommand{\PAProbeMae}{0.429}
\newcommand{\PAProbeRho}{0.342}
\newcommand{\PAProbeSlope}{0.143}
\newcommand{\PAProbeConstMae}{0.449}
\newcommand{\PAProbeUnitMae}{0.354}
\newcommand{\PAProbeUnitRho}{0.489}
\newcommand{\PAProbeDensRhoTile}{0.569}
\newcommand{\PAProbeDensRhoUnit}{0.783}
\newcommand{\PAProbeCeilMae}{0.361}
\newcommand{\PAProbeCeilFrac}{0.228}
\newcommand{\PAProbeRandRho}{0.341}

\newcommand{\PAProbeResK}{64}
\newcommand{\PAProbeResDim}{3584}

\newcommand{\PAProbeResConstMae}{0.449}
\newcommand{\PAProbeResPooledMae}{0.428}
\newcommand{\PAProbeResPooledRho}{0.306}
\newcommand{\PAProbeResPooledSlope}{0.126}

\newcommand{\MAEWacPsnr}{19.8}

\newcommand{\MAEPosGc}{2672}

\newcommand{\OBN}{500}
\newcommand{\OBPolyN}{93}
\newcommand{\OBCov}{0.726}
\newcommand{\OBHl}{0.018}
\newcommand{\OBTransferMae}{0.097}
\newcommand{\OBTransferExact}{0.916}
\newcommand{\OBGf}{0.435}
\newcommand{\OBNaming}{0.806}
\newcommand{\OBCrep}{0.012}
\newcommand{\OBDOne}{0.772}
\newcommand{\OBPolyMae}{0.416}

\newcommand{\OBPolyCov}{0.634}
\newcommand{\OBPolyClosedMae}{0.413}
\newcommand{\OBOracleMae}{0.195}

\newcommand{\OBConstMae}{0.414}
\newcommand{\OBConstPrior}{3.5}
\newcommand{\OBStratRightMae}{0.087}
\newcommand{\OBStratRightN}{73}
\newcommand{\OBStratWrongMae}{1.38}
\newcommand{\OBStratWrongN}{20}
\newcommand{\OBPertSlope}{0.711}
\newcommand{\OBPertR}{0.808}

\newcommand{\OBSwapSlope}{0.84}

\newcommand{\OBShuffle}{0.529}

\ExplSyntaxOn
\NewExpandableDocumentCommand{\pct}{m}{\fp_eval:n{ round( 100 * (#1), 1 ) }\%}
\ExplSyntaxOff
\newcommand{\GVage}{SelenoVLM\textsubscript{age}}
\newcommand{\GVchrono}{SelenoVLM\textsubscript{chrono}}
\newcommand{\ind}{\mathds{1}}       

\definecolor{color1}{HTML}{448AFF}  
\definecolor{color2}{HTML}{1565C0}  
\definecolor{color3}{HTML}{009688}  
\definecolor{color4}{HTML}{8BC34A}  
\definecolor{color5}{HTML}{FFC107}  
\definecolor{color6}{HTML}{FF9800}  
\definecolor{color7}{HTML}{F44336}  
\definecolor{color8}{HTML}{AD1457}  

\pgfplotsset{
	colormap={RdBu_r}{
		color=(color1!100),
		color=(color2!80),
		color=(color3!60),
		color=(color4!40),
		color=(color5!40),
		color=(color6!60),
		color=(color7!80),
		color=(color8!100)
	}
}

\newcommand{\token}[1]{%
	\tcbox[on line, size=fbox, boxsep=1pt, colback=gray!15, arc=1pt]{\ttfamily\strut #1}\hspace{2pt}%
}

\definecolor{stringteal}{RGB}{0, 80, 80}
\newcommand{\str}[1]{%
  \texttt{\color{stringteal}``#1"}%
}

\newcommand{\ci}[2]{{\footnotesize\textcolor{black!45}{[#1,\,#2]}}}
\newcommand{\cistrip}[3]{%
  \begin{tikzpicture}[x=1.6cm, baseline=-0.5ex, line cap=round]
    \draw[gray!30, line width=0.4pt] (0,0) -- (1,0);
    \draw[gray!30, line width=0.4pt] (0,0) ++(0,-1.4pt) -- ++(0,2.8pt);
    \draw[gray!30, line width=0.4pt] (1,0) ++(0,-1.4pt) -- ++(0,2.8pt);
    \draw[color2, line width=0.9pt] (#2,0) -- (#3,0);
    \draw[color2, line width=0.9pt] (#2,0) ++(0,-1.7pt) -- ++(0,3.4pt);
    \draw[color2, line width=0.9pt] (#3,0) ++(0,-1.7pt) -- ++(0,3.4pt);
    \fill[color1] (#1,0) circle (1.2pt);
  \end{tikzpicture}%
}

\hypersetup{
    colorlinks=true,
    linkcolor=blue,
	citecolor=blue,
    filecolor=magenta,
    urlcolor=cyan,
    pdftitle={Verifiably grounded machine interpretation of lunar geology},
    pdfpagemode=FullScreen,
}

\title{Verifiably grounded machine interpretation of lunar geology}

\newif\ifuniqueAffiliation
\uniqueAffiliationtrue

\ifuniqueAffiliation 
\author{ \href{https://orcid.org/0009-0008-3051-5976}{\includegraphics[scale=0.06]{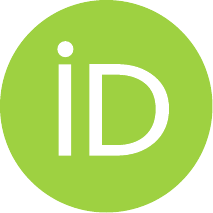}\hspace{1mm}\textcolor{black}{Tom Sander}}, \href{https://orcid.org/0000-0002-4324-1459}{\includegraphics[scale=0.06]{images/orcid.pdf}\hspace{1mm}\textcolor{black}{Kay Wohlfarth}}, \href{https://orcid.org/0000-0002-0715-0955}{\includegraphics[scale=0.06]{images/orcid.pdf}\hspace{1mm}\textcolor{black}{Christian Wöhler}} \\
	Image Analysis Group\\
	TU Dortmund University\\
	44227 Dortmund, Germany \\
	\texttt{\{firstname.lastname\}@tu-dortmund.de} \\
}
\else
\usepackage{authblk}

\newbox{\orcid}\sbox{\orcid}{\includegraphics[scale=0.06]{images/orcid.pdf}}
\author[1]{%
	\href{https://orcid.org/0000-0000-0000-0000}{\usebox{\orcid}\hspace{1mm}David S.~Hippocampus\thanks{\texttt{hippo@cs.cranberry-lemon.edu}}}%
}
\affil[1]{Department of Computer Science, Cranberry-Lemon University, Pittsburgh, PA 15213}
\fi

\renewcommand{\shorttitle}{Measured Grounding for a Lunar Vision-Language Model}

\begin{document}
\maketitle

\begin{abstract}
	Planetary geology relies on historical, interpretive reasoning to reconstruct past events from diverse observations. Here, we investigate how far this interpretive workflow can be automated by a multimodal vision-language model. Focusing on the stratigraphy of lunar basaltic mare volcanism, we train a model to generate verifiably grounded geologic interpretations directly from co-registered topographic, spectral, and geologic maps. We demonstrate that while the system successfully balances established geological priors with local visual evidence to accurately describe stratigraphy and terrain, numeric age dating derived solely from vision defaults to memorized priors. Integrating an open-book retrieval mechanism resolves this, enabling the model to faithfully cite published chronologies. Our findings delineate the necessary architecture for automated geologic inference: site evidence must be visually interpreted from local data, while quantitative historical context must be retrieved from the scientific record.
\end{abstract}


\section{Introduction}
\label{sec:introduction}

Planetary science attempts to explain the properties and evolutionary history of planetary bodies. To this end, a planetary scientist synthesizes a diverse set of observations, from remote sensing to in-situ measurements and laboratory analogues, into a coherent causal model of the body under study. The reasoning is interpretive and historical rather than experimental \citep{frodeman1995geological}, and it has often been compared to the work of Sherlock Holmes, who pieces together forensic clues to establish how past events unfolded \citep{anderl2016astronomy}. Once such a model is established, the scientist can interpret any specific region through the lens of the prevailing model and the local measurements.

Historically, this work has been carried out by trained experts with minimal automated support, generally restricted to terrain classifiers and classical data analysis. Machine intelligence has meanwhile reached near-expert performance in scientific and professional domains, including protein structure prediction \citep{jumper2021highly}, medium-range weather forecasting \citep{lam2023graphcast}, and legal examination \citep{katz2024gpt}. It may therefore now be possible to build a machine geologist. Because planetary data sources are inherently multimodal and most background knowledge exists as text and geologic maps, multimodal foundation models \citep{bommasani2021opportunities,MAE,alayrac2022flamingo,4M2024} offer a promising route to this goal.

While establishing new geologic models from data may be the prime goal of such a system, we start with the more constrained task of interpreting a given site under an established model and its local data. We focus on the stratigraphy of lunar basaltic mare volcanism, one of the best established models in planetary science \citep{head1992lunar,head1976lunar,wilson2017generation,wilhelms1971geologic,stoffler2001stratigraphy}. Its absolute chronology anchors relative crater size-frequency distributions to radiometrically dated Apollo and Luna samples \citep{neukum2001cratering}, and the resulting unit ages have been mapped across the nearside maria over more than two decades \citep{hiesinger2011ages,hiesinger2023lunar}. We develop an architecture that outputs the geologic interpretation of a given lunar region in textual form. The task poses several challenges, which we address in turn.

The first challenge is domain adaptation. General language models are pretrained on general-purpose world knowledge, and the scientific accuracy of their fluent statements about the Moon is unclear, so the model must be adapted to the lunar domain. Adaptation alone does not make the output trustworthy, however, since generative language models hallucinate \citep{maynez2020faithfulness}, which is unacceptable in a scientific instrument. The model must instead be grounded in the observational evidence of the presented site rather than in statistical correlations, and this grounding must be measured rather than asserted \citep{rashkin2023measuring}. Grounding in turn presupposes that the observations can enter the model at all. Vision-language models typically process 8-bit RGB images \citep{liu2023visual}, whereas digital elevation models, spectral ratio composites, and categorical geologic maps have no natural-image equivalent, so the architecture must ingest each modality while preserving its physical semantics rather than forcing it into a pseudo-RGB rendering.

Even a grounded and domain-adapted model faces a subtler failure mode. It can generate a plausible ``typical mare'' interpretation from literature priors alone, a subtle hallucination in which true general knowledge substitutes for site-specific evidence. Like a human geologist, the model must let priors constrain the hypothesis space while the local measurements decide, and we regard this balance as the computational core of interpretation itself. Where the background knowledge resides is itself a design question. It can live in the model's weights, instilled by domain adaptation, or in the context supplied at inference time \citep{lewis2020retrieval}. Closed-book reasoning resembles expert reasoning from memory but is opaque and hard to update, whereas open-book reasoning is traceable, citable, and updatable without retraining. Comparing both regimes is a controlled experiment on where scientific knowledge should reside in a machine geologist, and it ties into balancing, since internalized knowledge is exactly the strong prior that can drown out local evidence.

Here we show that the design of the supervision targets decides whether such a system is verifiably grounded, and that the same discipline exposes where grounding ends. Building on a multimodal masked autoencoder pretrained on co-registered lunar mosaics \citep{sander2026moon}, we bridge the frozen encoder into an adapted language model and train it on targets whose factual content cannot be reproduced without reading the presented tile. The generated interpretations depend causally on the tile, state raster-verifiable facts, and follow the stratigraphic model in where they speak about age. The stated numeric ages, however, track the global mare prior rather than the local evidence, although probes show that the visual representation carries a dating signal. In the open-book regime the same generation channel transfers published ages faithfully and causally from the retrieved record. The measurements delineate what a machine geologist reads from the site and what it must cite from the literature.

\section{Related Work}
\label{sec:related}

Lunar geology has long relied on painstaking manual mapping and crater counting, culminating in the interpretive framework we use today. The most widely accepted global product (the USGS Unified Geologic Map of the Moon \citep{fortezzo2020unified}) integrates Apollo-era 1:5{,}000{,}000 quadrangle maps. Their stratigraphic logic, established through nearside photogeologic mapping \citep{wilhelms1971geologic} and formalized in the time-stratigraphic systems of lunar history \citep{stoffler2001stratigraphy}, underpins our current understanding. Absolute ages for mare units rest on crater size-frequency distribution (CSFD) measurements calibrated against radiometrically dated Apollo and Luna samples \citep{neukum1984meteorite, neukum2001cratering, tanaka2012planetary}, a program that Hiesinger and colleagues have extended across the nearside maria over more than two decades \citep{hiesinger2000ages, hiesinger2003ages, hiesinger2010ages, hiesinger2011ages, hiesinger2023lunar}. These chronologies are the gold standard we validate against, but they are hand-made, spatially uneven, and in places decades old. They encode interpretation as static map units and tables rather than as a queryable, continuously updatable layer, and producing them does not scale to the global, high-resolution coverage that modern instruments now provide.

Machine learning has begun to automate the perceptual steps of planetary geology. Convolutional networks detect and measure impact craters directly from imagery and topography. The DeepMoon model, for instance, recovers crater catalogs from lunar digital elevation data and transfers to Mercury \citep{silburt2019lunar}, an approach since scaled to global multimodal crater catalogs \citep{la2025fromMoonToMercury} and to a global boulder map from LRO NAC imagery \citep{aussel2025global}, while semantic-segmentation approaches delineate terrain and geologic units for rover navigation and mapping \citep{kuang2022semantic}. More recently, an open-vocabulary detector built on a vision-language backbone has been fine-tuned to localize craters in high-resolution orbital imagery for landing-site assessment \citep{LunarCraterVLM}. Captioning systems such as SCOTI attach short textual labels to Martian terrain to prioritize downlink \citep{qiu2020scoti}. Each of these produces a perceptual product and stops there, whether a crater count, a label raster, or a brief caption. None of these systems reasons from the surface to its geoscientific interpretation, that is, to the inferred ages, the stratigraphy, and the formation history in which the actual science resides.

Recent advances in self-supervised masked modeling have shaped the learning of visual representations for planetary data. Masked autoencoders reconstruct missing image patches to learn transferable features \citep{MAE}, extended to multi-modal data by MultiMAE \citep{Multi-MAE} and generalized by the 4M framework for any-to-any prediction across modalities \citep{4M2024, 4M-21}, resting on discrete tokenization such as VQ-VAE \citep{VQVAE} and finite-scalar quantization \citep{mentzer2024fsq}. Prior work has applied this multimodal paradigm to lunar data, reconstructing imagery, topography, and albedo \citep{sander2026moon}. The focus of these methods remains perceptual reconstruction rather than language-based geological interpretation. A parallel effort has produced remote-sensing foundation models for Earth observation. SatMAE and Scale-MAE adapt masked autoencoding to the temporal, multispectral, and multiscale structure of satellite imagery \citep{SatMAE, reed2023scalemae}, and large pretrained backbones such as Prithvi and Clay, together with multimodal variants \citep{MMmesh, TerraMind}, serve as general feature extractors for downstream tasks \citep{Prithvi, claymodel}. These models are optimized for perceptual tasks on terrestrial sensors, producing embeddings or label maps rather than natural-language geological interpretation, and none targets a planetary body or its published geological record.

On the vision-language side, Frozen demonstrated the alignment of vision encoders with frozen language models for few-shot multimodal tasks \citep{tsimpoukelli2021multimodal}, Flamingo introduced gated cross-attention and Perceiver resampling to handle variable numbers of visual tokens \citep{alayrac2022flamingo, jaegle2021perceiver}, BLIP-2 and LLaVA established query-based bridging and visual instruction tuning \citep{li2023blip, liu2023visual}, and parameter-efficient adapters such as LoRA facilitate the language-model adaptation \citep{hu2022lora}. However, these methods usually rely on general web image-text data and lack mechanisms to ensure that the generated language is grounded in authoritative scientific sources. Retrieval-augmented generation has been proposed against this faithfulness gap, conditioning language models on retrieved documents to enable attribution of generated content to specific external corpora \citep{lewis2020retrieval}. This approach tackles the persistent problem of hallucination in generative models \citep{maynez2020faithfulness} and builds on methods that measure the attribution of statements to identifiable sources \citep{rashkin2023measuring}. However, existing studies in this area have focused almost exclusively on general-domain text and open-domain question answering, with limited or no application to the task of grounding geological interpretation in the primary planetary science literature.

Several recent systems relevant to geological interpretation are remote-sensing vision-language models. RemoteCLIP, GeoChat, RSGPT, and EarthGPT integrate visual encoders with language models to enable applications such as image captioning, visual question answering, and grounded dialogue over aerial and satellite imagery \citep{liu2024remoteclip, kuckreja2024geochat, hu2025rsgpt, zhang2024earthgpt}. Extensions of these systems have introduced retrieval mechanisms and agentic reasoning, supporting geological applications such as lithology recognition and mineral exploration \citep{chen2025vision, yu2025sta}. However, these models predominantly target terrestrial data and typically produce general scene descriptions. Even those oriented towards geology focus on surface material recognition, rather than reconstructing units' ages, stratigraphy, or formation histories, and none provide outputs grounded in a published, citable geological literature. Closest to our setting, LLaVA-LE adapts a LLaVA backbone to lunar imagery using model-written captions conditioned on gravity and slope rasters \citep{LunarLLaVA}. Its evaluation relies on language-model judges that score responses against caption-derived references without access to the image. Grounding is therefore asserted by the supervision and the protocol rather than measured. Neither the causal dependence of the text on the tile nor the faithfulness of stated quantities to the rasters is tested, and crater-chronology ages are not addressed. The task also differs from ours. LLaVA-LE answers concise questions about single-band panchromatic patches, whereas we target paragraph-length geological interpretation over co-registered multimodal tiles, designed so that the stated content can be verified against the input data. Since the model weights from \citet{LunarLLaVA} are not released, a comparison under a shared evaluation protocol is not possible.

\section{Methods}
\label{sec:method}

\begin{figure}
  \centering
  \begin{subfigure}[t]{\linewidth}
    \centering
    \begin{adjustbox}{max width=\linewidth}
      \input{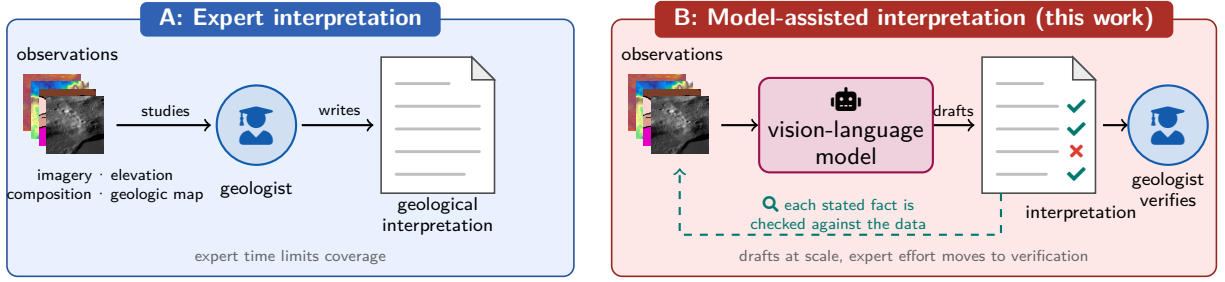}
    \end{adjustbox}
    \caption{The interpretation workflow this work targets.}
    \label{fig:system:workflow}
  \end{subfigure}\\[8pt]
  \begin{subfigure}[t]{\linewidth}
    \centering
    \begin{adjustbox}{max width=\linewidth}
      \input{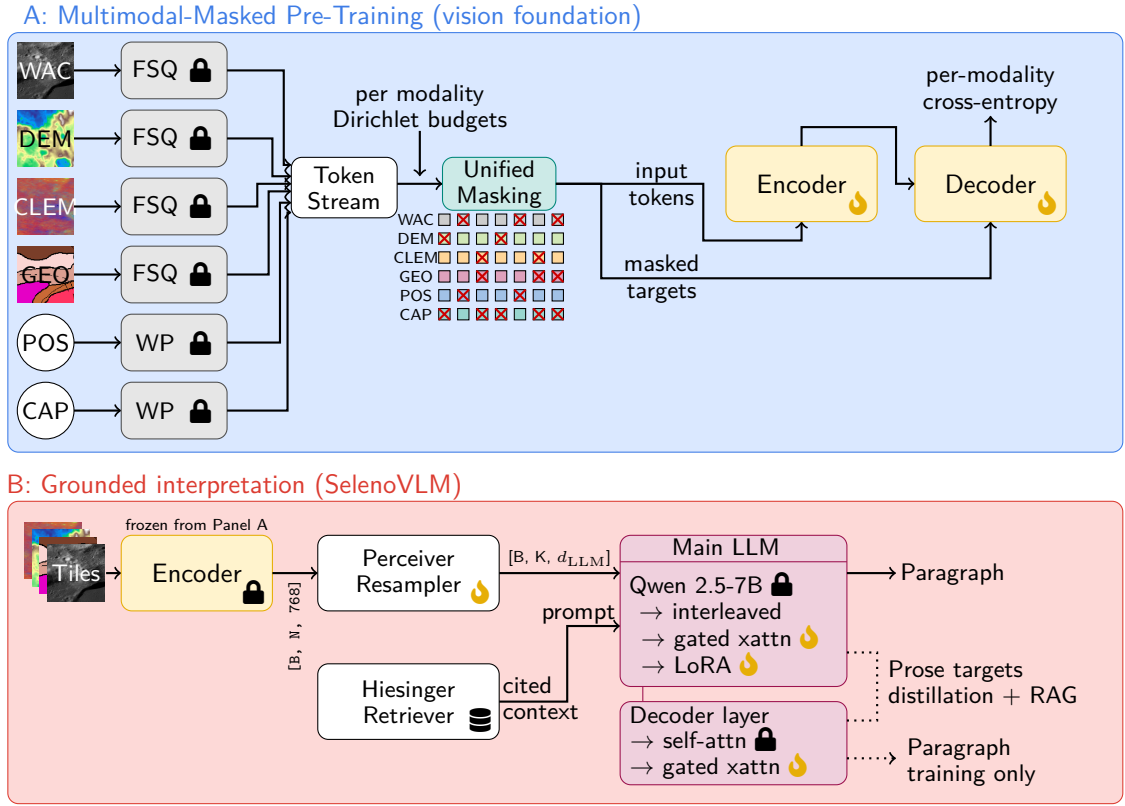}
    \end{adjustbox}
    \caption{The two-stage system.}
    \label{fig:system:overview}
  \end{subfigure}
  \caption{Overview. (a)~In established practice a geologist studies the co-registered observations and writes the geological interpretation, so coverage is limited by expert time. In this work a vision-language model drafts the interpretation from the same observations. Each stated fact is checked against the underlying rasters, and the expert verifies rather than transcribes. (b)~Stage A pretrains a multimodal masked autoencoder over the tokenized co-registered mosaics together with the coordinate string (POS) and the weakly supervised caption fields, collected in the CAP row (crossed cells denote masked tokens). Stage B freezes this encoder as a vision tower reading the raster modalities only and bridges it into a LoRA-adapted 7B language model through a Perceiver resampler and $\tanh$-gated cross-attention, supervised by distilled prose targets that open with deterministic raster-derived facts (Sec.~\ref{sec:method:targets}). \faLock{} marks frozen components and \faFire{} trainable ones.}
  \label{fig:system}
\end{figure}

The methodology consists of a two-stage process, as depicted in Fig.~\ref{fig:system}b. The initial stage (Sec.~\ref{sec:method:stageA}) uses a multimodal masked autoencoder (MAE) based on the 4M architecture~\citep{4M2024} for self-supervised vision representation learning. Each co-registered lunar modality is tokenized independently, and a unified encoder--decoder transformer is trained to reconstruct masked tokens from the visible tokens of the other modalities. In the subsequent stage (Sec.~\ref{sec:method:stageB}), the pre-trained and frozen MAE encoder serves as a vision tower and is integrated with a large language model via a lightweight, trainable bridge, forming a vision--language model (SelenoVLM) that generates natural-language geological interpretations grounded in retrieved domain literature. We note that the weakly supervised captions of Sec.~\ref{sec:data:text} enter the first stage as masked-reconstruction modalities, whereas the second stage uses them only in the distilled form of Sec.~\ref{sec:method:targets}. The MAE is trained once and then applied read-only during the second stage. Sec.~\ref{sec:method:targets} specifies the construction of the supervision targets, which carries the main methodological weight of this work, and Secs.~\ref{sec:eval:split}--\ref{sec:eval:vlm} define the evaluation protocol under which all results are scored.

\subsection{Data sources}
\label{sec:data}
\label{sec:data:sources}
All modalities are derived from globally co-registered Lunar Reconnaissance Orbiter (LRO) \citep{LRO1, LRO2} products and ancillary datasets. These datasets are resampled to a unified equirectangular grid at a nominal \SI{100}{\metre\per pixel} (Sec.~\ref{sec:data:tiling}) to enable pixel-level fusion. The dataset includes a panchromatic image (WAC) \citep{wagner2015newWAC}, elevation data (GLD100) \citep{scholten2012gld100}, a Clementine UVVIS color-ratio composite (Clementine) \citep{eliason1999clementine,hare2008clementine}, segmentation data (geomap) \citep{fortezzo2020unified}, and five text modalities, as summarized in Table~\ref{tab:data:modalities} and Fig.~\ref{fig:data:inputs} (Appendix~\ref{app:data}). Although the framework can be extended to incorporate additional co-registered rasters, such as the Moon Mineralogy Mapper (M$^3$) hyperspectral cube~\citep{m3}, only the listed modalities are utilized in this study.

\begin{table}[htbp]
    \centering
    \caption{Modalities used in this work. All raster products are co-registered at \SI{100}{\metre\per pixel}. The text modalities are derived from per-tile coordinates and from weakly supervised captions (Sec.~\ref{sec:data:text}).}
    \label{tab:data:modalities}
    \begin{tabular}{@{}llp{0.46\linewidth}@{}}
        \toprule
        Modality & Type & Source / derivation \\
        \midrule
        WAC                 & Image (1-ch)        & LROC Wide-Angle Camera global mosaic, \SI{100}{\metre}, June 2013~\citep{speyerer2011lunarWAC,wagner2015newWAC} \\
        Elevation (GLD100)  & Image (1-ch)        & LRO/LOLA-derived digital elevation model (eight equatorial PDS tiles, stitched)~\citep{scholten2012gld100} \\
        Clementine          & Image (3-ch)        & Clementine UVVIS color-ratio composite, warped to ULCN 2005~\citep{eliason1999clementine,hare2008clementine} \\
        Geomap              & Segmentation (50)   & Unified lunar geological map with categorical legend~\citep{fortezzo2020unified} \\
        POS                 & Text                & Per-tile geographic coordinates \\
        Primary Units       & Text                & Caption section \\
        Dominant Processes  & Text                & Caption section \\
        Stratigraphy        & Text                & Caption section \\
        Interpretation      & Text                & Caption section \\
        \bottomrule
    \end{tabular}
\end{table}

\subsection{Tiling and data splitting}
\label{sec:data:tiling}
Following alignment, the global mosaics are partitioned into square tiles with side length $S=512$ pixels. Tiles are extracted using a sliding window (stride $\tau=128$ pixels), producing spatially overlapping samples and thereby increasing the dataset size. Each tile spans approximately \SI{51.2}{\kilo\metre} east-west at the equator and \SI{67.4}{\kilo\metre} north-south, since the mosaic grid is anisotropic (303.2 and 230.3 pixels per degree of longitude and latitude, respectively). The overlap enlarges the dataset but creates a leakage risk, hence a spatially disjoint block split replaces randomized partitioning \citep{roberts2017crossValidation} and removes adjacency between tiles in different partitions. When elevation data are included, all modalities are cropped to the valid range of $[-79^{\circ}, 79^{\circ}]$ latitude.

The global grid is subdivided into macro-blocks, each deterministically assigned to one partition, and only tiles fully contained within a single partition are retained, which ensures zero pixel overlap between partitions. Applied within the captioned region used in this study, the split diminishes spatial autocorrelation and yields $169{,}784$ partitioned tiles, of which $132{,}462$ ($78.0\%$) form the training, $21{,}714$ ($12.8\%$) the validation, and $15{,}608$ ($9.2\%$) the test partition. The block size, the hash-based assignment, and the boundary handling are given in Appendix~\ref{app:data}. In addition to its raster form, the geological map is converted from its RGB legend into per-pixel unit labels spanning $50$ units, including background. The legend provides abbreviations and descriptions for each unit, which later form part of the retrieval corpus.

\subsection{Stage A: multimodal masked pre-training}
\label{sec:method:stageA}
\label{sec:data:text}
\label{sec:method:tokenizer}
Textual supervision at this stage is machine-generated. Each tile carries its central coordinates as a degrees-minutes-seconds string (the POS modality), and a vision-capable language model (\texttt{Gemma-3})~\citep{gemma3} writes structured descriptions for each captioned tile in four sections (Primary Units, Dominant Processes, Stratigraphy, Interpretation Justification), each of which forms one text modality. These weak labels are noisy and occasionally hallucinated, which the retrieval-grounded distillation of Sec.~\ref{sec:method:stageB} addresses. Tiles lacking captions inherit the caption of the spatially nearest annotated tile. This inheritance affects only training-time text, since at evaluation every stated fact is scored against the held-out tile's own rasters and every age against the withheld dated record (Sec.~\ref{sec:eval:vlm}).

All modalities are mapped to discrete token sequences so that a single transformer operates on a homogeneous stream. Each raster modality is tokenized by its own pre-trained and subsequently frozen tokenizer, a Vision Transformer encoder~\citep{ViT} discretized by finite scalar quantization (FSQ)~\citep{mentzer2024fsq}, with a conditional diffusion decoder~\citep{ho2020denoising} reconstructing the raster from the quantized codes. POS is tokenized at the character level and the caption fields with word-level WordPiece vocabularies~\citep{schuster2012japanese,wu2016google}. The codebook constructions, the vocabulary sizes, and tokenized examples are given in Appendix~\ref{app:tokenizers}.

At each training step, the tokenized modalities are concatenated into one stream and split into encoder (visible) and decoder (target) sets, with per-modality token budgets sampled from a Dirichlet mixture under fixed budgets \citep{Multi-MAE}, random masking for rasters~\citep{MAE} and span masking for text. With a fixed probability, one raster modality is withheld from the encoder entirely and used as a full decoder target, which places cross-modal reconstruction explicitly into the training distribution. The architecture is a standard transformer encoder--decoder with per-modality embedding tables and output heads over a shared embedding space, trained with per-token cross-entropy computed separately per modality. At inference, raster targets are decoded MaskGIT-style~\citep{chang2022maskgit} and text targets greedily autoregressive. Widths, depths, and the masking schedule are listed in Table~\ref{tab:impl-stageA} (Appendix~\ref{app:training}).

\subsection{Stage B: grounded interpretation (SelenoVLM)}
\label{sec:method:stageB}
In the second stage, language generation capabilities are integrated with the frozen vision foundation through a compact, trainable bridge, following the approach established by Flamingo~\citep{alayrac2022flamingo}. For each tile, the four raster modalities are provided to the frozen MAE encoder as complete (unmasked) inputs, producing a latent sequence and a corresponding key-padding mask. The text modalities are deliberately withheld from the encoder in this stage. The caption fields are the material from which the distilled targets are built, so encoding them would leak the answer into the vision path, and the coordinate string would reintroduce the location shortcut discussed in Sec.~\ref{sec:method:targets}. The weakly supervised structured captions are likewise not used directly as targets in this stage.

To bridge the vision and language components, a Perceiver resampler~\citep{jaegle2021perceiver} compresses the variable-length encoder output into $K=64$ learned query tokens projected to the language model's hidden width. These visual tokens are injected into a pre-trained Qwen2.5-7B-Instruct decoder~\citep{qwen25} via gated cross-attention blocks inserted before every fourth decoder layer, and LoRA adapters~\citep{hu2022lora} on the attention and feed-forward projections adapt the language model. Each block carries a $\tanh$ gate on its attention path, initialized at $0.1$ rather than $0$, which exposes the language model to the visual pathway from the first optimization step. The vision-free feed-forward sublayer of the block is disabled, hence any contribution of the bridge passes through cross-attention. Only the resampler, the cross-attention blocks, the adapters, and the auxiliary heads described below are updated during training.

The supervision targets are pre-distilled offline so that they integrate retrieved factual information rather than reproducing synthetic bullet-point captions. For each captioned tile, a language model (Gemma-3-27B) rewrites the four structured caption fields and retrieved domain context into an integrated natural-language paragraph, which is stored as a per-tile sidecar and retrieved at training time by position (Sec.~\ref{sec:method:targets}). Domain context is supplied by a retriever operating over a corpus of Hiesinger geological results~\citep{hiesinger2000ages,hiesinger2010ages,hiesinger2011ages,hiesinger2023lunar}, the book ``New views of the moon 2'' \citep{neal2023newViewsOfMoon2}, the geological-map unit descriptions, and the named features of the IAU gazetteer \citep{iau_usgs_2026}. During both training and inference, the retriever parses tile coordinates and returns a concise, referenced context block with the nearest dated unit within an $8^{\circ}$ geographic gate, the relevant map-unit description, and the most similar literature excerpts. Since the distilled targets integrate these facts, retrieval provides a substantive gradient signal during training. The corpus composition and indexing are detailed in Appendix~\ref{app:corpus-chrono}.

Model training uses a standard language modeling cross-entropy loss, computed solely on the answer tokens, with the system prompt and retrieved context excluded from the loss. The deterministic fact and age tokens of Sec.~\ref{sec:method:targets} are upweighted sixfold in this loss. Two auxiliary objectives act on the mean-pooled resampler output during training only. A unit head (weight $1.0$) classifies the $50$ geologic-map units, which grounds the resampler categorically without teacher forcing, and a vision-consistency contrastive loss (weight $0.5$, margin $1.0$) requires the answer to score better under its own tile's visual tokens than under those of another tile. Later variants additionally train a diagnostic age regression head on the same pooled vector, whose behavior is analyzed in Sec.~\ref{sec:res-gap}. Detailed optimization settings are provided in Appendix~\ref{app:training}.

Checkpoints are snapshotted densely during training, and the served checkpoint is selected at the knee of the trade-off between causal grounding and target adherence (Sec.~\ref{sec:res-targets}), never by validation loss. This selection was consistent across seven consecutive training runs, and a third epoch degraded the model in three independent attempts.

\subsection{Target construction}
\label{sec:method:targets}
The supervision targets, not the architecture, carry the grounding claim of this paper, so we state their construction in full. Each target consists of a deterministic opener with tile-specific numeric facts, followed on mare tiles by one age sentence, and then the distilled interpretive paragraph of Sec.~\ref{sec:method:stageB}. The governing rule is that a target must not be reproducible without reading the tile.

The served design (\GVage{}) opens every target with four facts computed from the rasters. Albedo is the tile mean of the normalized WAC reflectance, stated to two decimals. Relief is the spread between the 2nd and the 98th percentile of the tile's elevation values, which is insensitive to single outlier pixels, rounded to the nearest \SI{100}{\metre}. Roughness is the standard deviation of the elevation, rounded to the nearest \SI{25}{\metre}. The spectral fact is the tile mean of the Clementine 415/750 ratio, stated to two decimals. Each opener is phrased through one of four rotated sentence scaffolds, so that no single surface string can be memorized, and every scaffold carries an ordinal quintile descriptor (such as \str{very dark}) together with the rounded number. The number is the load-bearing element. With ordinal wording alone, only $35\%$ of intra-mare targets remained distinct. The rounded numbers at the served granularity raise this fraction to $44\%$, and coarsening the meter bins to \SI{250}{\metre} and \SI{50}{\metre} collapses it to $13\%$, which motivated keeping the fine bins (a second, terrain-adaptive coarsening is falsified in Sec.~\ref{sec:res-targets}). The scoring tolerances of Sec.~\ref{sec:eval:vlm} are matched to this granularity and enforced identically between the target builder and the scorer.

Age language is gated on the mapped geology. A tile receives an age sentence only if at least half of its geologic-map pixels belong to mare units, the same criterion by which the evaluation partitions tiles into mare and highland. If the nearest dated mare unit lies within $8^{\circ}$ great-circle distance, the sentence states that unit's published crater-count model age in Ga, again through rotated scaffolds. Beyond this gate the sentence names only the epoch. Every mare tile hence carries age language in one of the two forms, and highland tiles carry none, which is the behavior the emission metrics of Sec.~\ref{sec:res-emission} measure.

The crater-chronology variant (\GVchrono{}, Sec.~\ref{sec:res-agevalues}) extends this design as far as we could push it on the target side. Every target opens additionally with the tile's crater density from the Robbins database~\citep{robbins2019new}, counted for craters of at least \SI{1}{\kilo\metre} diameter per \SI{1000}{\kilo\metre\squared} and quantized in steps of $1$ below a density of $25$ and steps of $5$ above. This density is the quantity on which crater chronology rests, related to the absolute model age through the lunar chronology function of \citet{neukum2001cratering} (Appendix~\ref{app:corpus-chrono}). Mare tiles inside a dated mapped polygon state that unit's published age, determined by a point-in-polygon test on the full mapped geometry. This test replaces the nearest-centroid label, which deviates from the polygon age by $0.19$\,Ga on average even for tiles inside a polygon. Mare tiles outside every dated polygon state an age fitted from their own measured crater density through an isotonic regression trained on the in-polygon tiles, which calibrates the raw tile densities to the published record, so that all age sentences of this design are chronology-consistent with one another.

The assembled targets are stored per tile ($174{,}035$ rows for the served design) and retrieved by nearest position within a $3^{\circ}$ gate, with a deterministic prose template as fallback where no stored target exists. Teacher forcing hence pairs a held-out tile with the target built from its own rasters rather than a neighbor's. Prompts and targets carry no coordinates at any point. In an early bridge run with raw coordinates in the prompt, the language model learned a coordinate-to-geology shortcut and the cross-attention gates remained near zero, hence location enters only internally through the retriever.

\subsection{Test split and evaluation discipline}
\label{sec:eval}
\label{sec:eval:split}
All quantitative claims in this paper are computed on the test split of the geographic block split, which is spatially disjoint from the training data. The vision tower is scored on 240 seeded tiles of this split, and the bridge on $n=\GVN{}$ tiles. Paired comparisons, that is, the instructed-baseline comparison and the target-design ablation, use an identical seeded subset of $n=\BLGemmaFourN{}$ tiles so that per-tile differences are meaningful. Targets and closed-book prompts carry no coordinates at any point. The open-book context of Sec.~\ref{sec:res-openbook} is the one exception, since a retrieved record quotes the location of its published unit.

Where we report uncertainty, we use $95\%$ bootstrap confidence intervals over the evaluation tiles ($10{,}000$ resamples). Adjacent tiles within a partition overlap through the sliding window, so the scored tiles are spatially autocorrelated. The $\GVN{}$ scored tiles fall into $34$ distinct macro-blocks, and we therefore resample entire macro-blocks (a cluster bootstrap) rather than individual tiles. The cluster intervals are materially wider than their naive per-tile counterparts, and every significance statement in this paper rests on the cluster intervals.

\subsection{Vision-tower metrics}
\label{sec:eval:mae}
The frozen MAE is evaluated in two regimes. In the cross-modal generation regime, the target modality is fully withheld from the encoder and regenerated from all remaining modalities, with MaskGIT-style \citep{chang2022maskgit} iterative decoding for raster and segmentation targets and greedy autoregressive decoding for text. In the masked-reconstruction regime, training-style Dirichlet span masking is applied and token metrics are computed on the masked target positions only. Fidelity is measured in task space, with PSNR and SSIM for rasters \citep{Wang2004SSIM}, pixel accuracy and mIoU for the geologic map, BLEU-4 \citep{papineni2002bleu} and ROUGE-L \citep{lin2004rouge,ganesan2018rouge} for the caption sections, and parse rate together with great-circle error for position. Teacher-forced top-1 accuracy and perplexity are additionally reported as token-level calibration, with the uniform floor of each vocabulary as the chance reference (Table~\ref{tab:mae-tower}).

\subsection{SelenoVLM metrics}
\label{sec:eval:vlm}
Let $\mathcal{T}$ denote the set of held-out tiles. For a tile $t \in \mathcal{T}$, let $v_t$ be its visual tokens, $y_t$ the free-running generation, and $y_t^{\ast} = (y_{t,1}^{\ast}, \dots, y_{t,L_t}^{\ast})$ the teacher-forced target sequence. The bridge is scored in five metric groups, mirroring the blocks of Table~\ref{tab:main-results}.

The first group measures causal grounding through interventions on the visual input, in the spirit of decision-flip tests \citep{serrano2019attentionInterpretability} and modality-ablation analyses for multimodal models \citep{gat2021perceptualScore,frank-etal-2021-vision}. Writing $p_\theta(\cdot \mid y^{\ast}_{t,<i}, v)$ for the next-token distribution under visual input $v$, the tile-swap flip rate is the fraction of target positions whose argmax changes when the tile's visual tokens are replaced by those of a different tile $t' \neq t$,
\begin{equation}
  f_{\mathrm{swap}} = \frac{1}{|\mathcal{T}|} \sum_{t \in \mathcal{T}} \frac{1}{L_t} \sum_{i=1}^{L_t}
  \ind\!\left[ \arg\max_{w} p_\theta(w \mid y^{\ast}_{t,<i}, v_{t'}) \neq \arg\max_{w} p_\theta(w \mid y^{\ast}_{t,<i}, v_{t}) \right],
  \label{eq:flip}
\end{equation}
and the vision-removal flip rate $f_{\mathrm{zero}}$ is defined analogously with $v_{t'}$ replaced by zeroed visual latents. A model that uses vision only in an on/off manner shows a large $f_{\mathrm{zero}}$ together with $f_{\mathrm{swap}} \approx 0$, whereas a tile-grounded model must flip under both interventions. At the generation level, we report the mean token-set Jaccard distance $1 - |W_t \cap W_t^{0}| / |W_t \cup W_t^{0}|$ between paired generations with and without vision \citep{jaccard1901etude,levandowsky1971distance}, and unit naming, the fraction of tiles where $y_t$ contains a content word from the mapped dominant unit's legend description.

Grounding alone does not guarantee that the stated content is correct, so the second group scores it against the rasters. Every target opens with deterministic facts computed from the tile (Sec.~\ref{sec:method:targets}). For each fact $k$ (albedo, relief, roughness, 415/750 ratio), let $x_{t,k}$ be the raster-measured value, $q_k(x_{t,k})$ its quantization to the target granularity, and $\hat{x}_{t,k}$ the value parsed back out of $y_t$ (absent facts are recorded as unstated). A stated fact counts as correct if
\begin{equation}
  \bigl| \hat{x}_{t,k} - q_k(x_{t,k}) \bigr| \leq \Delta_k,
  \label{eq:facttol}
\end{equation}
where $\Delta_k$ is a fixed tolerance matched to the target granularity, that is, $\Delta = 0.03$ for albedo, $100\,$m for relief, $25\,$m for roughness, and $0.05$ for the ratio, enforced identically between the target builder and the scorer. Stated-fact fidelity is the mean per-fact accuracy over the stated facts, and per-quantity accuracy and MAE are reported alongside the aggregate.

For ages, we separate the decision to state an age from the stated value. The stated age $\hat{a}_t = \operatorname{median}\{a : \text{``}a\,\text{Ga''} \in y_t\}$ is the median of all Ga-qualified numbers in the generation, which is insensitive to an isolated anomalous value, and it remains undefined if no age is mentioned. A tile counts as mare if at least half of its geologic-map pixels belong to mare units (Sec.~\ref{sec:method:targets}), and as highland otherwise. Mare coverage is the fraction of mare tiles with $\hat{a}_t$ defined, and the highland false-positive rate is the same fraction over highland tiles, where no basalt age is defined and silence is the correct behavior. Epoch scoring maps the named epoch word, or otherwise the stated age, onto the conventional model-age bounds \citep{stoffler2001stratigraphy} and requires exact agreement with the reference age's epoch, with an adjacent relaxation of one epoch.

The value metric $\hat{a}_t$ is evaluated independently. Unless a comparison is explicitly declared open book, all age evaluations are closed-book, that is, the dated reference record that indicates the tile's location is omitted from the retrieved context, so any stated age must be inferred rather than copied. For tiles that state an age, we report the mean absolute error (MAE) $\frac{1}{n}\sum_t |\hat{a}_t - a_t|$ and the Spearman rank correlation. Where the scatter is analyzed (Table~\ref{tab:age-poly}), we additionally report the least-squares slope of the stated on the reference ages (one under perfect extraction, zero for a constant predictor) and the signed mean bias. The same definitions apply to the probe predictions of Sec.~\ref{sec:res-gap}, with $\hat{a}_t$ taken as the probe output. Every age comparison keeps a constant-prior baseline visible, the constant predictor $\tilde{a} = \operatorname{median}\{a_t\}$ scored on the identical tiles, which states the global mare prior everywhere. A model without visual age extraction matches this baseline, hence recovery claims must beat it. We note that the constant-prior MAE differs numerically across tables, since it is recomputed on each analysis's own held-out set, and it is always the fair within-table reference. Two ground-truth conventions appear in this paper. Centroid ages assign a tile the published age of the nearest dated unit within the $8^{\circ}$ retrieval gate ($n=\GVN{}$), and polygon-accurate ages come from point-in-polygon tests on the full mapped geometry ($n=93$). The two are never mixed within a comparison. The probe battery that reads the frozen representations under a leave-unit-out protocol is specified in Appendix~\ref{app:probes}.

Sec.~\ref{sec:res-openbook} additionally evaluates the deployed open-book mode, in which the retriever supplies its full context including the nearest dated record within the $8^{\circ}$ gate. The injected centroid record is identical to the centroid ground truth, so agreement with it is circular by construction. We therefore report it as transfer fidelity, the mean absolute difference between the stated and the injected age together with the fraction of exact transfers (within $0.05$\,Ga), and never as age recovery. Non-circular open-book accuracy is scored against polygon ground truth only, where the injected record can disagree with the point-in-polygon age, and every open-book age comparison keeps two references visible, the constant prior and a retrieval oracle that scores the injected record itself against the polygon ground truth. Two evaluation-time interventions, a $\pm 0.5$\,Ga rewrite of the injected age and a swap to a different tile's full context, together with a geo-suppressed control that removes only the dated record, make the integration claim causal rather than correlational.

The last group tracks degeneration in free-running generation. Distinct-1 \citep{li2016diversity} is the mean ratio of unique to total tokens per generation, averaged over tiles rather than normalized over a whole corpus of generations as in the original definition, and the consecutive-repeat rate is the mean fraction of adjacent token pairs that are identical. We treat these statistics as a first-class selection axis, since repeat rates above a few percent collapse the generated output at later checkpoints.

We note that perplexity (ppl), token accuracy, and ROUGE against the distilled targets are fit diagnostics on circular supervision. They guide training but support no grounding claim, and no such claim in this paper rests on them. The instructed baselines span open-weights models (gemma4:26b \citep{gemma4}, qwen3.5:27b \citep{qwen35}, deepseek-vl2 \citep{deepseekvl2}) and closed frontier systems (gpt-5.6-terra \citep{gptterra}, claude-opus-5 \citep{claudeopus5}).

All baselines receive the same four co-registered rasters rendered as images together with an explicit task instruction and are scored by the identical parser and protocol. A prose-fallback parser ensures they are not penalized on output format. All models in this comparison, including \GVage{}, generate under a uniform answer budget of 1200 new tokens, which no completion exhausts, so no score is confounded by truncation. Two decoding deviations remain, the reasoning setting of gpt-5.6-terra and the sampling of claude-opus-5, and are stated in the caption of Table~\ref{tab:baselines}.

\section{Results}
\label{sec:results}
This section follows the workflow of the interpreting geologist, from reading the tile, through interpretation within the stratigraphic model of mare volcanism, to dating by visual assessment and from the published record. All results are computed on the spatially disjoint test split (Sec.~\ref{sec:eval:split}), and all metrics are those of Sec.~\ref{sec:eval:vlm}. Age metrics are closed book, with the location's dated record withheld from the prompt, except in the open-book evaluation of Sec.~\ref{sec:res-openbook}.

\subsection{A verifiably grounded interpreter of lunar terrain}
\label{sec:res-grounding}
Interpretation begins with the examination of observations, and the introduction required this grounding to be measured rather than asserted. The tile-swap ablation of Sec.~\ref{sec:eval:vlm} provides this measurement. Substituting the visual tokens of the input tile with those of another alters the argmax at \pct{\GVShuffle} of target positions, close to the \pct{\GVZero} when vision is removed altogether, and the stated facts verify against the rasters. The strongest instructed baseline trails in stated-fact fidelity by a factor of 1.6 under an identical protocol.

Three properties of the frozen vision tower constrain what a tile can convey and underpin all language-side claims (Table~\ref{tab:mae-tower}). First, the withheld camera mosaic is reconstructed from elevation, geologic map, and spectral composite at \MAEWacPsnr\,dB PSNR, so the spectral basalt evidence needed for mare-age analysis is recoverable from appearance. Second, structure is largely predictable from the co-registered context whereas tile-specific wording is not, so a target without tile-specific facts can be reproduced without reading the tile. Third, tiles are not geolocatable. Coordinates generated from imagery alone carry a mean great-circle error of \MAEPosGc\,km, within 2\% of the random-pair expectation, so the grounding measured below cannot rest on coordinate memorization.

\begin{table}[htbp]
  \centering
  \caption{Held-out evaluation of the stage-A multimodal MAE on the spatially disjoint test split ($n=240$ seeded tiles), before any language coupling. In the cross-modal regime, the target modality is fully withheld from the encoder and regenerated from the remaining modalities (MaskGIT for rasters, greedy autoregression for text). Fidelity is reported in task space because FSQ codes are degenerate (exact token agreement is near zero while decoded rasters remain faithful). Top-1 and perplexity are teacher-forced token calibration. Chance ppl is the uniform floor of each vocabulary. The reconstruction column uses training-style random masking. A high position error is the desired outcome. A single tile is not geolocatable, hence downstream grounding cannot be coordinate memorization.}
  \label{tab:mae-tower}
  \adjustbox{max width=\linewidth}{%
  \begin{tabular}{llcccc}
    \toprule
    Modality & Cross-modal generation & top-1 $\uparrow$ & ppl $\downarrow$ & chance ppl
      & recon top-1 $\uparrow$ \\
    \midrule
    WAC & PSNR 19.8\,dB / SSIM 0.53 & 0.01 & 3091.1 & 6561 & 0.01 \\
    Elevation & PSNR 7.2\,dB / SSIM 0.35 & 0.01 & 2392.5 & 6561 & 0.02 \\
    Clementine & PSNR 15.0\,dB / SSIM 0.46 & 0.02 & 1129.7 & 15360 & 0.04 \\
    Geomap & pixel acc.\ 0.30 / mIoU 0.09 & 0.12 & 222.9 & 2916 & 0.67 \\
    Position & parse 0.96 / GC err.\ 2672\,km & 0.87 & 1.4 & 65 & 0.82 \\
    Units & BLEU-4 0.08 / ROUGE-L 0.15 & 0.93 & 1.2 & 295 & 0.80 \\
    Processes & BLEU-4 0.17 / ROUGE-L 0.29 & 0.90 & 1.4 & 841 & 0.80 \\
    Stratigraphy & BLEU-4 0.02 / ROUGE-L 0.11 & 0.79 & 2.1 & 3111 & 0.68 \\
    Interpretation & BLEU-4 0.04 / ROUGE-L 0.14 & 0.69 & 2.8 & 4176 & 0.63 \\
    \bottomrule
  \end{tabular}%
  }
\end{table}

On this foundation, the closeness of the two ablation rates is the important observation (Table~\ref{tab:main-results}). Earlier bridges on near-identical targets showed the inverse signature, with vision as a mere on/off cue. The present model produces text that depends on which tile it sees and hence describes the actual site rather than generic mare terrain.

Beyond being tile-specific, the stated content verifies against the rasters. The model names the mapped geologic unit on \pct{\GVNaming} of held-out tiles, and stated-fact fidelity reaches \GVGf{} in aggregate (\GVGfMare{} on mare, \GVGfHighland{} on highland tiles). Mean albedo is stated within one quantization level on \pct{\GVAlbAcc} of tiles and the Clementine 415/750 ratio on \pct{\GVRatAcc}, whereas relief and roughness are harder (\pct{\GVRelAcc} and \pct{\GVRghAcc}), with errors concentrated in high-relief highland regions. Generation remains clean at the served checkpoint (Table~\ref{tab:main-results}), and the same reading sustains multi-turn dialogue, in which the model derives the superposition order of the units (Fig.~\ref{fig:chat-example}).

\begin{figure}[htbp]
  \centering
  \begin{adjustbox}{max width=\linewidth, max totalheight=0.75\textheight}
    \input{figures/results/fig_chat_example.tex}
  \end{adjustbox}
  \caption{Closed-book multi-turn chat on a held-out test tile in Mare Serenitatis
    (16.7\textdegree{}E, 22.3\textdegree{}N). A Copernican crater unit (Cc, bright ejecta)
    is superposed on Eratosthenian mare basalts (Em). The model reads the four co-registered
    rasters only. Neither coordinates nor retrieval context enter the prompt. Turn 1 states the
    raster facts (albedo and 415/750 ratio within one scoring tolerance, relief and roughness
    underestimated) and an age of 3.7\,Ga against a published unit age of 3.5\,Ga. The
    follow-up turns answer without restating the opener (cross-turn 4-gram overlap $<0.02$) and
    derive the superposition order that ends with the cross-cutting crater unit. Each turn is
    the best of 10--12 sampled candidates, selected for raster grounding (turn 1) and
    non-repetition (follow-ups). Named features quoted by the model (e.g.\ Oceanus Procellarum)
    are decorative and frequently wrong under the closed-book protocol, and they are not
    verified against the gazetteer. The answers are abridged at the marked ellipses.}
  \label{fig:chat-example}
\end{figure}

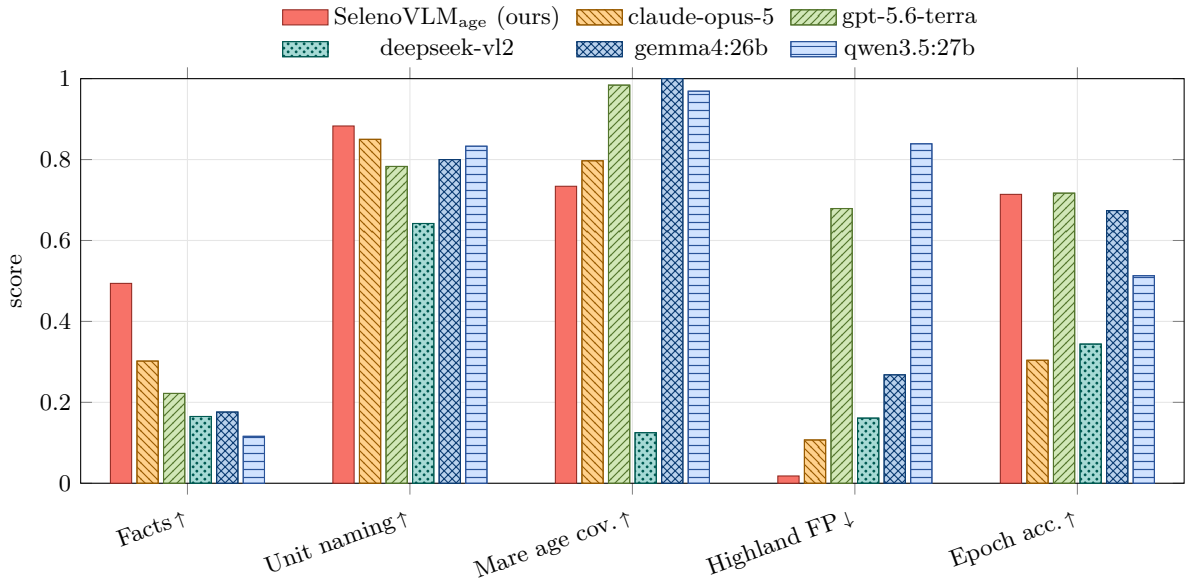
\begin{figure}
  \centering
\usetikzlibrary{patterns}
\begin{tikzpicture}
  \begin{axis}[
      width=0.98\linewidth, height=0.42\linewidth,
      ybar, bar width=8pt, ymin=0, ymax=1.0,
      enlarge x limits=0.12,
      symbolic x coords={Facts,Unit naming,Mare age cov.,Highland FP,Epoch acc.}, xtick=data,
      xticklabels={%
        Facts\,{\scriptsize$\uparrow$},%
        Unit naming\,{\scriptsize$\uparrow$},%
        Mare age cov.\,{\scriptsize$\uparrow$},%
        Highland FP\,{\scriptsize$\downarrow$},%
        Epoch acc.\,{\scriptsize$\uparrow$}},
      x tick label style={font=\footnotesize, rotate=20, anchor=north east},
      ylabel={score}, label style={font=\small},
      y tick label style={font=\footnotesize},
      legend style={font=\footnotesize, at={(0.5,1.02)},
      anchor=south, legend columns=3, draw=none, fill=none,
      /tikz/every even column/.append style={column sep=4pt}},
      area legend,
      grid=major, grid style={black!10},
  ]
    \addplot[
      draw=color7!60!black,
      fill=color7!75!white,
    ] coordinates { (Facts,0.494) (Unit naming,0.883) (Mare age cov.,0.734) (Highland FP,0.018) (Epoch acc.,0.714) };
    \addlegendentry{\GVage{} (ours)}
    \addplot[
      draw=color6!60!black,
      fill=color6!45!white,
      postaction={pattern=north west lines, pattern color=color6!60!black}
    ] coordinates { (Facts,0.302) (Unit naming,0.850) (Mare age cov.,0.797) (Highland FP,0.107) (Epoch acc.,0.304) };
    \addlegendentry{claude-opus-5}
    \addplot[
      draw=color4!60!black,
      fill=color4!40!white,
      postaction={pattern=north east lines, pattern color=color4!60!black}
    ] coordinates { (Facts,0.222) (Unit naming,0.783) (Mare age cov.,0.984) (Highland FP,0.679) (Epoch acc.,0.717)  };
    \addlegendentry{gpt-5.6-terra}
    \addplot[
      draw=color3!60!black,
      fill=color3!35!white,
      postaction={pattern=crosshatch dots, pattern color=color3!60!black}
    ] coordinates { (Facts,0.165) (Unit naming,0.642) (Mare age cov.,0.125) (Highland FP,0.161) (Epoch acc.,0.344) };
    \addlegendentry{deepseek-vl2}
    \addplot[
      draw=color2!60!black,
      fill=color2!30!white,
      postaction={pattern=crosshatch, pattern color=color2!60!black}
    ] coordinates { (Facts,0.176) (Unit naming,0.800) (Mare age cov.,1.000) (Highland FP,0.268) (Epoch acc.,0.674) };
    \addlegendentry{gemma4:26b}
    \addplot[
      draw=color1!60!black,
      fill=color1!25!white,
      postaction={pattern=horizontal lines, pattern color=color1!60!black}
    ] coordinates { (Facts,0.116) (Unit naming,0.833) (Mare age cov.,0.969) (Highland FP,0.839) (Epoch acc.,0.513) };
    \addlegendentry{qwen3.5:27b}
  \end{axis}
\end{tikzpicture}
  \caption{Instructed open-weights and frontier (V)LM baselines vs.\ \GVage{} on identical tiles, protocol, and scoring ($n=\BLGemmaFourN$). All models generate under an identical 1200-token answer budget, which no model exhausts. The full metric set, including the constant-age-prior comparison, appears in Table~\ref{tab:baselines}. The baselines read the photograph-like quantities at best moderately and the physically calibrated ones poorly, and none combines mare age coverage with highland silence. The low highland false-positive rate of deepseek-vl2 reflects near-global age silence (mare age coverage \BLDeepseekvlTwoCov{}) rather than a learned gate. gpt-5.6-terra nominally exceeds \GVage{} on epoch accuracy (\BLGptFiveSixterraEpAcc{} against 0.714) while stating mare ages on \pct{\BLGptFiveSixterraHl} of highland tiles, where no basalt age is defined. Arrows on the category labels mark whether a higher ($\uparrow$) or lower ($\downarrow$) score is better, the highland false-positive rate being the only category where lower is better.}
  \label{fig:baselines}
\end{figure}

Instructed vision-language models do not attain comparable grounding under the same protocol and scoring criteria (Fig.~\ref{fig:baselines}, Table~\ref{tab:baselines}, Sec.~\ref{sec:eval:vlm}). All models processed identical rendered rasters on a seeded subset of $n=\BLGemmaFourN$ tiles. \GVage{} achieved a stated-fact fidelity of \ABIncGf{} on this subset, the strongest baseline, claude-opus-5 \citep{claudeopus5}, attained \BLClaudeopusFiveGf{}, and the lower $95\%$ confidence bound for \GVage{} (0.440) surpasses every baseline point estimate.
The strongest baselines are substantially larger models, so the gap cannot be explained by model capacity, and it concentrates on the data types without a natural-image analog, on the physically calibrated Clementine 415/750 ratio in particular (Table~\ref{tab:baselines}). No baseline simultaneously achieves mare age coverage and highland silence (Sec.~\ref{sec:res-emission}), nor does any outperform its constant-age prior for stated age values (Sec.~\ref{sec:res-agevalues}). We note that this comparison conflates domain training, raster rendering, and target supervision, hence the attribution of the grounding to the target design rests on the single-variable falsifications of Sec.~\ref{sec:res-targets}.

\begin{table}[htbp]
  \centering
  \caption{Comparison against instructed open-weights and frontier (V)LMs on identical held-out tiles, protocol, and scoring ($n=\BLGemmaFourN$). Baselines receive the four co-registered rasters rendered as images plus an explicit task instruction, and all age metrics are closed-book. All models generate under an identical 1200-token answer budget, which no model exhausts, since every completion terminates itself before the cap. The gpt-5.6-terra run used a low reasoning-effort setting with reasoning enabled, whereas thinking was off or unavailable for the other baselines, and the claude-opus-5 run is not sampled at temperature 0, since the API rejects sampling parameters (Sec.~\ref{sec:eval:vlm}). Bold marks the best value per column. $^*$Mare age coverage below 0.15. The age cells of this row rest on a handful of emissions, hence they are not comparable and are excluded from the per-column bolding. $^{\dagger}$On the full centroid-GT set (Table~\ref{tab:main-results}) the served model nominally edges the constant baseline by 0.005\,Ga, but the bootstrapped difference straddles zero and the leave-one-region-out control (Table~\ref{tab:loo}) attributes the remainder to memorized regional structure, hence we score it as not beating the prior.}
  \label{tab:baselines}
  \adjustbox{max width=\linewidth}{
  \begin{tabular}{lcccccccc}
    \toprule
    Model & Facts $\uparrow$ & Albedo $\uparrow$ & 415/750 $\uparrow$ & Unit naming $\uparrow$
      & Mare age cov.\ $\uparrow$ & Highland FP $\downarrow$ & Epoch acc.\ $\uparrow$
      & Beats const.? \\
    \midrule
    \GVage{} (served) & \textbf{0.494} & \textbf{0.567} & \textbf{0.433} & \textbf{0.883} & 0.734 & \textbf{0.018} & 0.714 & no$^{\dagger}$ \\
    \midrule
    gemma4:26b & \BLGemmaFourGf & \BLGemmaFourAlbAcc & \BLGemmaFourRatAcc & \BLGemmaFourNaming & \textbf{\BLGemmaFourCov} & \BLGemmaFourHl & \BLGemmaFourEpAcc & \BLGemmaFourBeats \\
    qwen3.5:27b & \BLQwenThreeFiveGf & \BLQwenThreeFiveAlbAcc & \BLQwenThreeFiveRatAcc & \BLQwenThreeFiveNaming & \BLQwenThreeFiveCov & \BLQwenThreeFiveHl & \BLQwenThreeFiveEpAcc & \BLQwenThreeFiveBeats \\
    deepseek-vl2 & \BLDeepseekvlTwoGf & \BLDeepseekvlTwoAlbAcc & \BLDeepseekvlTwoRatAcc & \BLDeepseekvlTwoNaming & \BLDeepseekvlTwoCov$^*$ & \BLDeepseekvlTwoHl & \BLDeepseekvlTwoEpAcc & \BLDeepseekvlTwoBeats \\
    \midrule
    gpt-5.6-terra & \BLGptFiveSixterraGf & \BLGptFiveSixterraAlbAcc & \BLGptFiveSixterraRatAcc & \BLGptFiveSixterraNaming & \BLGptFiveSixterraCov & \BLGptFiveSixterraHl & \textbf{\BLGptFiveSixterraEpAcc} & \BLGptFiveSixterraBeats \\
    claude-opus-5 & \BLClaudeopusFiveGf & \BLClaudeopusFiveAlbAcc & \BLClaudeopusFiveRatAcc & \BLClaudeopusFiveNaming & \BLClaudeopusFiveCov & \BLClaudeopusFiveHl & \BLClaudeopusFiveEpAcc & \BLClaudeopusFiveBeats \\
    \bottomrule
  \end{tabular}
  }
\end{table}

\begin{table}
  \centering
  \caption{Held-out evaluation of \GVage{} on the spatially disjoint test split ($n=\GVN$ tiles). All age metrics are closed-book, i.e.\ the location's dated reference is withheld from the prompt. Stated-fact tolerances are matched to the target quantization ($\pm$0.03 albedo, $\pm$100\,m relief, $\pm$25\,m roughness, $\pm$0.05 ratio). Brackets give the $95\%$ cluster-bootstrap confidence interval over the 34 macro-blocks of the test split (Sec.~\ref{sec:eval:split}), shown again as a strip on a $[0,1]$ axis with the dot at the point estimate. Rows without an interval are point estimates, and the age-value difference, which lies off this axis, is reported numerically only. The age-value block is computed on the 172 tiles with a stated age and a defined reference, and its nominal skill is analyzed in Sec.~\ref{sec:res-agevalues} and Table~\ref{tab:loo}.}
  \label{tab:main-results}
  \adjustbox{max width=\linewidth}{%
  \begin{tabular}{llr l c}
    \toprule
    & Metric & Value & \multicolumn{2}{c}{$95\%$ CI} \\
    \midrule
    \multirow{4}{*}{Vision grounding}
    & tile-swap flip rate (shuffle) $\uparrow$        & 0.520 & \ci{0.481}{0.562} & \cistrip{0.520}{0.481}{0.562} \\
    & vision-removal flip rate (zero) $\uparrow$      & 0.631 & \ci{0.522}{0.740} & \cistrip{0.631}{0.522}{0.740} \\
    & names mapped geologic unit $\uparrow$           & 0.858 & & \\
    & generation change, vision ablated $\uparrow$    & 0.824 & & \\
    \midrule
    \multirow{4}{*}{Stated-fact fidelity}
    & overall                                         & 0.458 & \ci{0.400}{0.511} & \cistrip{0.458}{0.400}{0.511} \\
    & mare / highland                                 & 0.527 / 0.399 & & \\
    & albedo acc.\ / MAE                              & 0.516 / 0.043 & & \\
    & 415/750 ratio acc.\ / MAE                       & 0.369 / 0.091 & & \\
    \midrule
    \multirow{4}{*}{Age emission}
    & mare coverage $\uparrow$ ($n=230$)              & 0.730 & \ci{0.597}{0.837} & \cistrip{0.730}{0.597}{0.837} \\
    & mare age MAE (Ga) $\downarrow$                  & 0.426 & & \\
    & highland false-positive rate $\downarrow$ ($n=270$) & 0.015 & \ci{0.000}{0.034} & \cistrip{0.015}{0.000}{0.034} \\
    & epoch accuracy / adjacent $\uparrow$            & 0.659 / 1.000 & & \\
    \midrule
    \multirow{3}{*}{Age-value skill}
    & model MAE vs.\ constant-age baseline (Ga)       & 0.425 vs.\ 0.430 & & \\
    & difference model $-$ constant (Ga)              & $-0.005$ & \ci{-0.022}{+0.014} & \\
    & Spearman $\rho$ (stated vs.\ reference)         & 0.295 & & \\
    \midrule
    \multirow{2}{*}{Degeneration}
    & consecutive-repeat rate $\downarrow$            & 0.0039 & & \\
    & distinct-1 $\uparrow$                           & 0.780 & & \\
    \bottomrule
  \end{tabular}%
  }
\end{table}

\subsection{Grounding is created by the targets and decays with training}
\label{sec:res-targets}
\begin{figure}[htbp]
  \centering
  \begin{tikzpicture}
  \begin{axis}[
      width=0.95\linewidth, height=0.4\linewidth,
      xlabel={training step}, ylabel={score},
      label style={font=\small}, tick label style={font=\footnotesize},
      ymin=0, ymax=1.0, xmin=0,
      legend pos=south east, legend cell align=left,
      legend style={font=\footnotesize, draw=none, fill=none},
      grid=both, grid style={black!10},
  ]
    \addplot[thick, mark=*, mark size=1.2pt, color1!75!black] coordinates { (2000,0.341) (4000,0.281) (6000,0.491) (8000,0.579) (8279,0.542) (8400,0.607) (10000,0.586) (12000,0.652) (14000,0.622) (16000,0.632) (16400,0.588) };
    \addlegendentry{causal grounding (tile-swap flip)}
    \addplot[thick, mark=square*, mark size=1.2pt, color3!90!black] coordinates { (2000,0.459) (4000,0.275) (6000,0.450) (8000,0.450) (8279,0.419) (8400,0.387) (10000,0.419) (12000,0.425) (14000,0.302) (16000,0.401) (16400,0.414) };
    \addlegendentry{stated-fact fidelity}
    \addplot[thick, mark=triangle*, mark size=1.5pt, color5!80!black] coordinates { (2000,0.619) (4000,0.286) (6000,0.714) (8000,0.619) (8279,0.762) (8400,0.286) (10000,0.667) (12000,0.524) (14000,0.238) (16000,0.238) (16400,0.381) };
    \addlegendentry{mare age coverage (closed-book)}
    \draw[dashed, thick, black!60] (axis cs:10000,0) -- (axis cs:10000,1)
      node[pos=0.97, anchor=north east, font=\footnotesize]{selected checkpoint};
  \end{axis}
\end{tikzpicture}
  \caption{Training dynamics of the bridge. Causal vision grounding (tile-swap flip) trends upward over training, while adherence to target content (stated facts, closed-book age coverage) peaks early and decays. The served checkpoint is selected at the knee of this trade-off, never by validation loss, which measures fit to distilled targets rather than grounding. Late checkpoints additionally degenerate into repetition (not shown).}
  \label{fig:training-curve}
\end{figure}
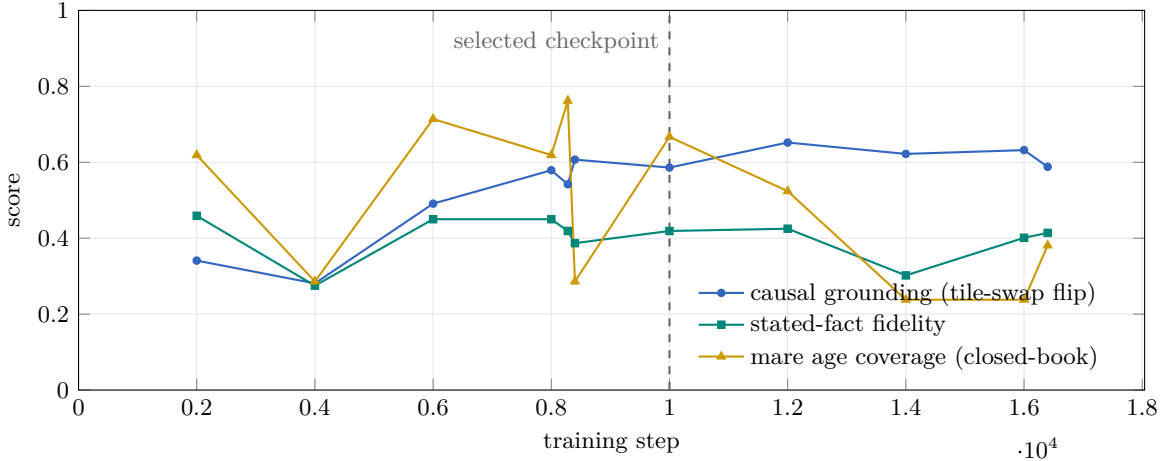

Two observations locate this grounding in the target design rather than in the architecture. Fitting the targets competes with reading the tile over training, and coarsening the injected facts lifts content metrics at the cost of causal grounding. The first observation is the training dynamics (Fig.~\ref{fig:training-curve}). Causal grounding, measured by the tile-swap flip, rises over training apart from local fluctuations, whereas alignment with the target content peaks in the first epoch and decays. Validation loss, which quantifies fit to the distilled targets, selects a strictly worse checkpoint in every run, and the served checkpoint is chosen at the knee of this trade-off (Sec.~\ref{sec:method:stageB}).

The second observation concerns the granularity of the injected facts (Table~\ref{tab:bins-ablation}). Terrain-adaptive meter bins, which coarsen the relief and roughness targets on high-relief tiles, make highland quantities more learnable and lift every content metric, yet the paired tile-swap flip drops by $0.093 \pm 0.024$ ($t=-3.90$, $n=120$). Coarser targets are satisfiable with less tile-specific vision, so target learnability trades against grounding, and the served design keeps the fine bins (Sec.~\ref{sec:method:targets}).

\begin{table}[htbp]
  \centering
  \caption{Terrain-adaptive meter-fact bins vs.\ the served fixed fine bins, retrained and evaluated on the identical $n=120$ seeded held-out tiles. Adaptive bins make highland quantities learnable (highland fact fidelity 0.375\,$\to$\,0.398) and lift mare fidelity, unit naming, and age coverage, but causal grounding regresses (paired tile-swap flip $\Delta = -0.093 \pm 0.024$, $t = -3.90$), since coarser targets are satisfiable with less tile-specific vision. The fixed-bin model remains the served checkpoint.}
  \label{tab:bins-ablation}
  \adjustbox{max width=\linewidth}{
  \begin{tabular}{lcccccccc}
    \toprule
    Targets & Facts $\uparrow$ & Facts (mare) & Facts (highl.) & Unit naming $\uparrow$
      & Mare age cov.\ $\uparrow$ & Tile-swap flip $\uparrow$ & Consec.\ rep.\ $\downarrow$
      & Distinct-1 $\uparrow$ \\
    \midrule
    Fixed fine bins (served) & 0.494 & 0.598 & 0.375 & 0.883
      & 0.734 & 0.539 & 0.004 & 0.776 \\
    Terrain-adaptive bins & 0.523 & 0.633 & 0.398 & 0.900
      & 0.750 & 0.445 & 0.008 & 0.767 \\
    \midrule
    Paired $\Delta$ (adaptive $-$ fixed) & \multicolumn{4}{l}{facts: $-0.015 \pm 0.028$ ($t=-0.52$)}
      & & \multicolumn{3}{l}{tile-swap: $-0.093 \pm 0.024$ ($t=-3.90$)} \\
    \bottomrule
  \end{tabular}
  }
\end{table}

\subsection{Age behavior follows the stratigraphic model}
\label{sec:res-emission}
The stratigraphic model fixes where age statements are meaningful, on the basalts of the maria and not on the highlands, and the learned emission behavior follows this structure. \GVage{} provides an age for \pct{\GVCov} of mare tiles ($n=\GVMareN$), while remaining nearly silent on highland terrain, with a false-age rate of \pct{\GVHl} ($n=\GVHlN$). A stated epoch is correct in \pct{\GVEpAcc} of cases and adjacent in the remainder. The closest baseline, claude-opus-5, states false highland ages on \pct{\BLClaudeopusFiveHl} of highland tiles against 1.8\% for \GVage{} on the identical subset (Fig.~\ref{fig:baselines}).

Two observations clarify what governs this behavior. First, the model learns when to remain silent. With age sentences omitted from a quarter of mare targets, emission was unstable across checkpoints, but stabilized once every mare target included age language. Second, explicit prompting does not replace target design, since instructions to state a numeric age did not increase coverage. Emission is hence governed by target statistics rather than by inference-time instructions. It remains to examine whether the stated values are derived from the tile.

\subsection{Stated age values follow the prior rather than the local evidence}
\label{sec:res-agevalues}
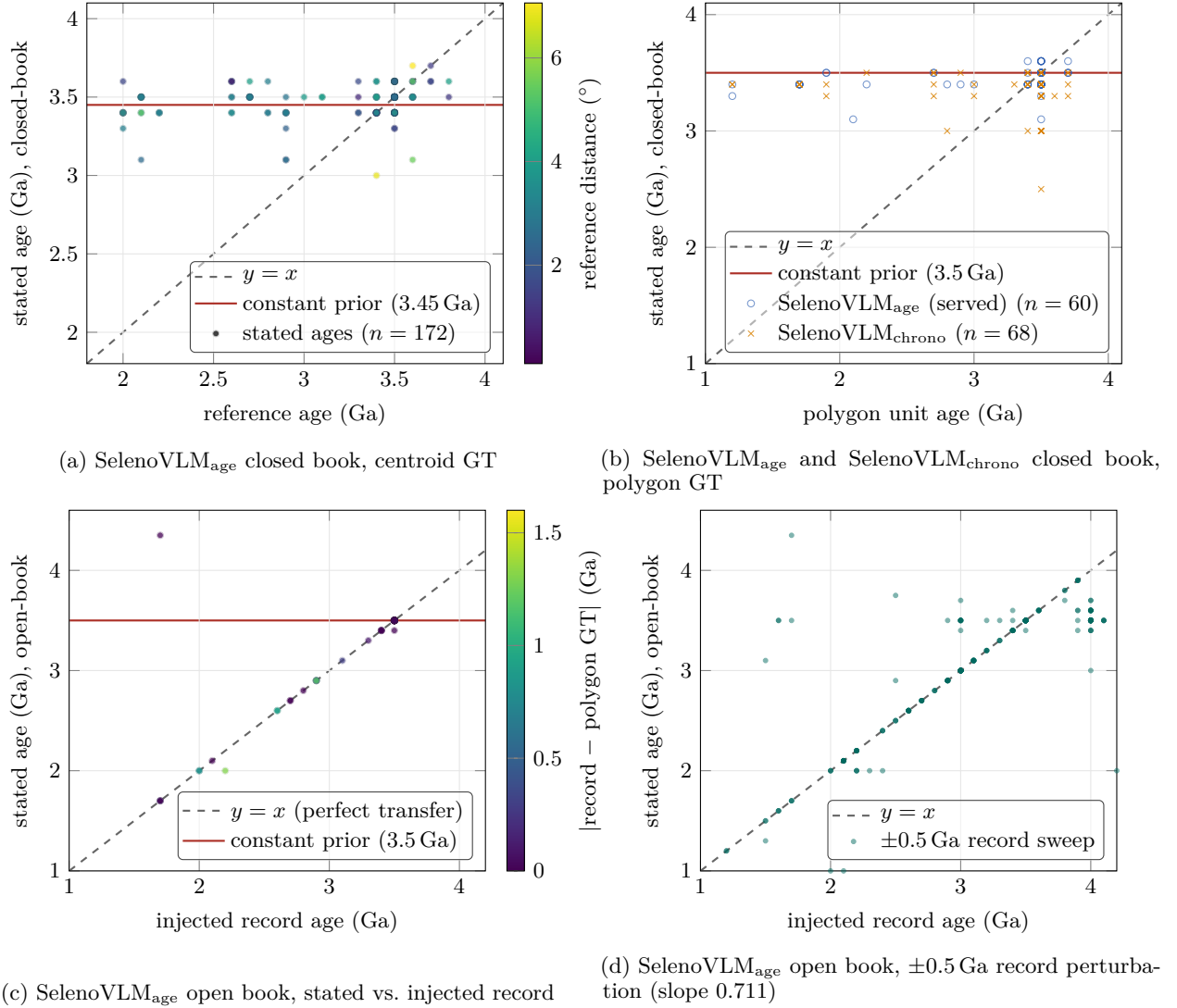
\begin{figure*}
  \centering
  \begin{subfigure}[t]{0.48\linewidth}
    \centering
    \begin{tikzpicture}
  \begin{axis}[
      width=0.95\linewidth, height=0.85\linewidth,
      xlabel={reference age (Ga)}, ylabel={stated age (Ga), closed-book},
      label style={font=\small}, tick label style={font=\footnotesize},
      xmin=1.8, xmax=4.1, ymin=1.8, ymax=4.1,
      colorbar, colorbar style={ylabel={reference distance ($^\circ$)}, width=2.5mm, ylabel style={font=\small}, tick label style={font=\footnotesize}},
      colormap name=viridis, legend pos=south east, legend cell align=left,
      legend style={font=\footnotesize, draw=white!25!black, fill=white, fill opacity=0.5, text opacity=1, draw opacity=1, rounded corners=2pt, inner sep=1.5pt},
      axis on top, grid=both, grid style={black!10},
  ]
    \addplot[dashed, thick, black!60, domain=1.8:4.1] {x};
    \addlegendentry{$y=x$}
    \addplot[color7!70!black, thick, domain=1.8:4.1] {3.45};
    \addlegendentry{constant prior (3.45\,Ga)}
    \addplot[scatter, only marks, mark=*, mark size=1.3pt, scatter src=explicit,
              fill opacity=0.75, draw opacity=0.25]
      table[x index=0, y index=1, meta index=2, row sep=newline] {
    3.40 3.40 5.09
    3.40 3.50 1.81
    3.50 3.50 2.91
    3.50 3.40 0.65
    2.10 3.50 3.77
    3.50 3.40 2.50
    2.70 3.50 2.03
    3.50 3.50 3.47
    3.50 3.40 1.51
    3.50 3.40 3.74
    2.70 3.50 2.68
    3.40 3.40 0.77
    3.50 3.50 1.67
    2.90 3.40 1.07
    3.50 3.50 2.72
    3.50 3.50 2.34
    3.40 3.60 4.12
    3.50 3.40 3.42
    2.70 3.50 2.77
    3.50 3.40 3.51
    3.50 3.40 5.33
    3.50 3.40 4.73
    3.40 3.40 2.01
    2.10 3.10 2.55
    2.10 3.40 3.21
    3.40 3.40 0.78
    3.50 3.50 1.40
    3.50 3.50 2.66
    2.60 3.50 1.73
    3.50 3.60 3.21
    1.70 3.40 2.20
    3.50 3.50 2.95
    3.50 3.50 0.20
    2.00 3.40 2.66
    2.60 3.60 1.78
    3.80 3.60 4.00
    3.50 3.40 2.54
    3.50 3.50 1.29
    2.90 3.40 2.59
    3.60 3.50 2.15
    2.00 3.60 1.61
    3.50 3.40 5.04
    1.70 3.40 2.66
    2.00 3.40 0.77
    3.50 3.30 1.41
    3.40 3.40 2.54
    3.50 3.50 1.78
    3.40 3.40 1.01
    2.70 3.50 2.39
    3.50 3.50 1.28
    2.10 3.50 1.88
    3.40 3.60 3.49
    2.10 3.50 2.37
    3.50 3.50 2.79
    3.30 3.40 1.09
    2.10 3.50 4.65
    3.30 3.60 1.73
    2.90 3.10 1.46
    3.00 3.50 3.89
    3.50 3.50 0.77
    2.80 3.40 2.74
    3.10 3.50 4.19
    3.60 3.50 1.53
    3.50 3.40 4.33
    3.50 3.40 3.10
    3.50 3.60 1.85
    3.50 3.50 2.56
    1.70 3.40 1.99
    2.70 3.50 3.52
    3.50 3.60 2.17
    2.90 3.40 3.01
    3.50 3.60 3.35
    3.50 3.50 3.85
    2.20 3.40 4.21
    1.70 3.40 2.01
    3.50 3.60 5.85
    3.50 3.40 1.41
    3.50 3.50 2.30
    3.60 3.50 2.47
    3.60 3.50 2.76
    2.00 3.30 3.08
    3.50 3.60 0.50
    3.50 3.40 3.46
    3.50 3.50 1.32
    3.50 3.40 5.29
    2.60 3.60 2.28
    3.40 3.50 2.79
    2.10 3.50 1.54
    3.50 3.50 1.92
    2.60 3.50 3.51
    1.70 3.60 3.19
    3.50 3.60 3.00
    3.60 3.50 4.07
    2.70 3.60 3.98
    3.40 3.40 1.90
    3.50 3.50 0.90
    3.40 3.40 1.05
    2.90 3.30 2.30
    2.70 3.50 1.99
    3.70 3.60 1.83
    3.50 3.50 3.63
    3.40 3.50 3.60
    2.10 3.50 3.35
    3.30 3.40 2.31
    2.60 3.60 0.54
    3.50 3.40 1.47
    3.50 3.50 4.94
    2.60 3.40 3.29
    3.60 3.60 1.83
    3.40 3.00 6.83
    2.00 3.40 1.41
    2.90 3.10 2.87
    3.50 3.50 0.68
    3.60 3.60 2.22
    1.70 3.40 2.42
    2.80 3.40 3.07
    3.50 3.40 1.86
    3.50 3.40 2.51
    3.50 3.40 1.90
    3.50 3.60 2.03
    2.90 3.40 2.91
    3.60 3.60 5.33
    3.50 3.50 0.60
    3.40 3.40 0.25
    2.70 3.50 2.51
    3.50 3.60 3.05
    3.50 3.50 1.43
    3.40 3.50 3.20
    3.40 3.40 1.06
    2.00 3.40 5.12
    2.90 3.40 3.07
    1.70 3.40 2.46
    3.50 3.40 1.92
    3.50 3.40 1.93
    3.50 3.60 0.10
    3.40 3.40 1.51
    3.50 3.50 3.65
    3.30 3.50 0.93
    3.50 3.60 2.60
    3.50 3.50 1.36
    2.20 3.40 2.13
    3.50 3.60 5.12
    1.70 3.50 4.08
    3.40 3.40 3.14
    3.10 3.50 3.86
    2.90 3.50 2.79
    3.50 3.50 3.52
    2.70 3.50 3.30
    1.70 3.40 2.79
    3.50 3.50 1.92
    3.50 3.30 1.68
    3.50 3.40 1.59
    3.50 3.40 4.00
    2.60 3.40 3.44
    3.40 3.40 3.38
    2.10 3.40 5.29
    3.80 3.50 1.26
    3.50 3.60 1.32
    3.50 3.50 3.41
    2.20 3.40 3.30
    3.50 3.60 2.69
    3.60 3.70 7.06
    2.00 3.40 3.01
    3.60 3.50 3.37
    1.70 3.40 2.81
    2.80 3.60 3.07
    3.60 3.10 5.48
    3.70 3.60 1.53
    3.40 3.50 4.10
    2.10 3.50 3.07
    3.70 3.70 1.36
    3.50 3.50 2.37
    };
    \addlegendentry{stated ages ($n=172$)}
  \end{axis}
\end{tikzpicture}
    \vspace*{-1em}
    \caption{\GVage{} closed book, centroid GT}
    \label{fig:age-scatter}
  \end{subfigure}
  \hfill
  \begin{subfigure}[t]{0.48\linewidth}
    \centering
    \begin{tikzpicture}
  \begin{axis}[
      width=0.95\linewidth, height=0.85\linewidth,
      xlabel={polygon unit age (Ga)}, ylabel={stated age (Ga), closed-book},
      label style={font=\small}, tick label style={font=\footnotesize},
      xmin=1.0, xmax=4.1, ymin=1.0, ymax=4.1,
      legend pos=south east, legend cell align=left,
      legend style={font=\footnotesize, draw=white!25!black, fill=white, fill opacity=0.5, text opacity=1, draw opacity=1, rounded corners=2pt, inner sep=1.5pt},
      axis on top, grid=both, grid style={black!10},
  ]
    \addplot[dashed, thick, black!60, domain=1.0:4.1] {x};
    \addlegendentry{$y=x$}
    \addplot[color7!70!black, thick, domain=1.0:4.1] {3.5};
    \addlegendentry{constant prior (3.5\,Ga)}
    \addplot[only marks, mark=o, mark size=1.4pt, color1!70!black, opacity=0.65]
      table[x index=0, y index=1, row sep=newline] {
    3.50 3.40
    3.50 3.50
    3.50 3.50
    3.50 3.40
    1.70 3.40
    2.70 3.50
    1.90 3.50
    3.50 3.40
    3.50 3.40
    2.70 3.50
    3.40 3.40
    3.50 3.50
    2.90 3.40
    3.50 3.50
    3.50 3.40
    3.40 3.40
    2.10 3.10
    3.40 3.40
    3.50 3.50
    3.70 3.50
    3.50 3.50
    3.70 3.60
    2.20 3.40
    3.50 3.50
    3.50 3.50
    3.50 3.40
    3.70 3.50
    3.50 3.40
    3.50 3.40
    1.70 3.40
    1.20 3.40
    3.50 3.30
    3.40 3.40
    3.40 3.60
    3.70 3.50
    3.40 3.40
    3.50 3.10
    3.50 3.50
    2.80 3.40
    3.40 3.50
    3.50 3.40
    3.50 3.40
    3.50 3.60
    3.50 3.50
    1.70 3.40
    3.50 3.60
    1.70 3.40
    3.50 3.60
    1.90 3.50
    3.50 3.40
    1.70 3.40
    3.50 3.60
    3.50 3.40
    3.50 3.50
    1.20 3.30
    3.50 3.60
    3.00 3.40
    3.50 3.50
    3.50 3.40
    3.50 3.60
    };
    \addlegendentry{\GVage{} (served) ($n=60$)}
    \addplot[only marks, mark=x, mark size=1.7pt, color6!85!black, opacity=0.8]
      table[x index=0, y index=1, row sep=newline] {
    3.50 3.40
    3.50 3.00
    3.50 3.50
    3.50 3.40
    1.70 3.40
    2.70 3.30
    1.90 3.30
    3.50 3.40
    3.50 3.40
    2.70 3.50
    3.40 3.50
    3.40 3.00
    3.50 3.00
    2.90 3.50
    3.30 3.40
    3.50 3.00
    3.50 3.50
    3.50 3.40
    3.40 3.50
    3.50 3.30
    3.40 3.40
    3.50 3.30
    3.60 3.30
    3.70 3.50
    3.50 3.30
    3.70 3.50
    2.20 3.50
    3.50 3.30
    3.50 3.50
    3.00 3.30
    3.50 3.40
    3.70 3.40
    3.50 3.00
    3.50 3.40
    1.70 3.40
    1.20 3.40
    3.50 3.30
    3.40 3.40
    2.70 3.40
    3.40 3.40
    3.70 3.30
    3.40 3.40
    2.70 3.50
    3.50 3.30
    3.50 3.00
    2.80 3.00
    3.40 3.50
    3.50 3.40
    3.50 3.40
    3.50 3.30
    3.50 3.00
    1.70 3.40
    3.50 3.30
    1.70 3.40
    3.50 3.00
    1.90 3.40
    3.50 3.40
    1.70 3.40
    3.50 3.40
    3.50 3.00
    3.50 3.40
    3.50 3.00
    1.20 3.40
    3.50 3.40
    3.00 3.40
    3.50 2.50
    3.50 3.40
    3.50 3.50
    };
    \addlegendentry{\GVchrono{} ($n=68$)}
  \end{axis}
\end{tikzpicture}
    \caption{\GVage{} and \GVchrono{} closed book, polygon GT}
    \label{fig:age-poly}
  \end{subfigure}\\[4pt]
  \begin{subfigure}[t]{0.48\linewidth}
    \centering
\begin{tikzpicture}
  \begin{axis}[
      width=0.95\linewidth, height=0.85\linewidth,
      xlabel={injected record age (Ga)}, ylabel={stated age (Ga), open-book},
      label style={font=\small}, tick label style={font=\footnotesize},
      xmin=1.0, xmax=4.2, ymin=1.0, ymax=4.6,
      colorbar, colorbar style={ylabel={$|$record $-$ polygon GT$|$ (Ga)}, width=2.5mm, ylabel style={font=\small}, tick label style={font=\footnotesize}},
      colormap name=viridis, legend pos=south east, legend cell align=left,
      legend style={font=\footnotesize, draw=white!25!black, fill=white, fill opacity=0.5, text opacity=1, draw opacity=1, rounded corners=2pt, inner sep=1.5pt},
      axis on top, grid=both, grid style={black!10},
  ]
    \addplot[dashed, thick, black!60, domain=1.0:4.2] {x};
    \addlegendentry{$y=x$ (perfect transfer)}
    \addplot[color7!70!black, thick, domain=1.0:4.2] {3.5};
    \addlegendentry{constant prior (3.5\,Ga)}
    \addplot[scatter, only marks, mark=*, mark size=1.3pt, scatter src=explicit,
              fill opacity=0.75, draw opacity=0.25]
      table[x index=0, y index=1, meta index=2, row sep=newline] {
    3.40 3.40 0.10
    3.50 3.50 0.00
    3.50 3.50 0.00
    2.70 2.70 0.00
    3.50 3.50 1.60
    3.50 3.50 0.00
    3.50 3.50 0.00
    2.70 2.70 0.00
    3.40 3.40 0.00
    3.50 3.50 0.00
    2.90 2.90 0.00
    3.50 3.50 0.00
    3.50 3.50 0.00
    3.50 3.50 0.10
    2.10 2.10 0.00
    3.40 3.40 0.00
    3.50 3.50 0.00
    3.50 3.50 0.20
    2.60 2.60 0.90
    3.50 3.50 0.20
    1.70 1.70 0.50
    3.50 3.50 0.00
    3.50 3.50 0.00
    3.50 3.50 0.00
    3.50 3.50 0.20
    2.90 2.90 0.60
    3.50 3.40 0.00
    1.70 4.35 0.00
    2.00 2.00 0.80
    3.50 3.50 0.00
    3.40 3.40 0.00
    3.50 3.50 0.20
    3.30 3.30 0.10
    2.90 2.90 0.60
    3.50 3.50 0.00
    2.80 2.80 0.00
    3.10 3.10 0.30
    3.50 3.50 0.00
    3.50 3.50 0.00
    3.50 3.50 0.00
    3.50 3.50 0.00
    1.70 1.70 0.00
    3.50 3.50 0.00
    2.90 2.90 1.20
    3.50 3.50 0.00
    3.50 3.50 1.60
    2.20 2.00 1.30
    1.70 1.70 0.00
    3.50 3.50 0.00
    3.50 3.50 0.00
    3.50 3.50 0.00
    2.00 2.00 0.80
    3.50 3.50 0.00
    3.50 3.50 0.50
    3.50 3.50 0.00
    3.50 3.50 0.00
    2.60 2.60 0.90
    };
  \end{axis}
\end{tikzpicture}
    \caption{\GVage{} open book, stated vs.\ injected record}
    \label{fig:openbook-transfer}
  \end{subfigure}
  \hfill
  \begin{subfigure}[t]{0.48\linewidth}
    \centering
\begin{tikzpicture}
  \begin{axis}[
      width=0.95\linewidth, height=0.85\linewidth,
      xlabel={injected record age (Ga)}, ylabel={stated age (Ga), open-book},
      label style={font=\small}, tick label style={font=\footnotesize},
      xmin=1.0, xmax=4.2, ymin=1.0, ymax=4.6,
      legend pos=south east, legend cell align=left,
      legend style={font=\footnotesize, draw=white!25!black, fill=white, fill opacity=0.5, text opacity=1, draw opacity=1, rounded corners=2pt, inner sep=1.5pt},
      axis on top, grid=both, grid style={black!10},
  ]
    \addplot[dashed, thick, black!60, domain=1.0:4.2] {x};
    \addlegendentry{$y=x$}
    \addplot[only marks, mark=*, mark size=0.9pt, color3!70!black, opacity=0.5]
      table[x index=0, y index=1, row sep=newline] {
    2.90 2.90
    3.40 3.40
    2.90 2.90
    3.90 3.90
    3.50 3.50
    3.00 3.00
    4.00 3.50
    3.50 3.50
    3.00 3.00
    4.00 3.60
    2.10 2.10
    1.60 1.60
    2.60 2.60
    2.70 2.70
    2.20 2.20
    3.20 3.20
    3.50 3.50
    3.00 3.50
    4.00 3.50
    3.50 3.50
    3.00 3.00
    4.00 3.50
    3.50 3.50
    3.00 3.50
    4.00 3.50
    2.70 2.70
    2.20 2.20
    3.20 3.20
    3.40 3.40
    2.90 2.90
    3.90 3.50
    3.50 3.50
    3.00 3.50
    4.00 3.50
    2.90 2.90
    2.40 2.40
    3.40 3.60
    3.50 3.50
    3.00 3.00
    4.00 3.50
    3.50 3.50
    3.00 3.00
    4.00 3.50
    3.50 3.50
    3.00 3.00
    4.00 3.60
    3.40 3.40
    2.90 2.90
    3.90 3.90
    3.50 3.50
    3.00 3.00
    4.00 3.50
    2.70 2.70
    2.20 2.20
    3.20 3.20
    3.50 3.50
    3.00 3.00
    4.00 3.50
    3.50 3.50
    3.00 3.00
    4.00 3.50
    3.50 3.50
    3.00 3.70
    4.00 3.50
    3.40 3.40
    2.90 2.90
    3.90 3.60
    2.10 2.10
    1.60 3.50
    2.60 2.60
    2.10 1.00
    2.60 2.60
    4.00 3.50
    3.40 3.40
    2.90 2.90
    3.90 3.40
    3.50 3.50
    3.00 3.00
    4.00 3.50
    3.50 3.50
    3.00 3.00
    4.00 3.50
    2.60 2.60
    2.10 2.10
    3.10 3.10
    3.50 3.50
    3.00 3.00
    4.00 3.60
    1.70 1.70
    1.20 1.20
    2.20 2.20
    3.50 3.50
    3.00 3.00
    4.00 3.00
    3.50 3.50
    3.00 3.00
    4.00 3.50
    2.00 1.00
    1.50 1.50
    2.50 3.75
    2.60 2.60
    2.10 2.10
    3.10 3.10
    3.80 3.80
    3.30 3.30
    4.30 4.30
    3.50 3.50
    3.00 3.00
    4.00 3.60
    3.50 3.50
    3.00 3.00
    4.00 3.50
    2.90 2.90
    2.40 2.00
    3.40 3.50
    4.20 2.00
    3.60 3.60
    3.10 3.10
    4.10 3.50
    2.00 2.00
    1.50 1.50
    2.50 2.50
    3.50 3.40
    3.00 3.40
    4.00 3.40
    1.70 4.35
    2.20 2.00
    2.00 2.00
    1.50 1.30
    2.50 2.90
    3.50 3.50
    3.00 3.00
    4.00 3.70
    3.40 3.40
    2.90 2.90
    3.90 3.90
    3.50 3.50
    3.00 3.00
    4.00 3.50
    2.70 2.70
    2.20 2.20
    3.20 3.20
    3.50 3.50
    3.00 3.00
    4.00 3.50
    2.10 2.10
    1.60 3.50
    2.60 2.60
    3.40 3.40
    2.90 3.50
    3.90 3.60
    3.10 3.10
    2.10 2.10
    1.60 1.60
    2.60 2.60
    3.50 3.50
    3.00 3.00
    4.00 3.50
    3.30 3.30
    2.80 2.80
    3.80 3.70
    2.10 2.10
    1.60 1.60
    2.60 2.60
    3.40 3.40
    3.30 3.30
    2.80 2.80
    3.80 3.80
    2.90 2.90
    2.40 2.40
    3.40 3.40
    3.00 3.00
    2.50 2.50
    3.50 3.50
    3.50 3.50
    3.00 3.00
    4.00 3.60
    2.80 2.80
    2.30 2.00
    3.30 3.50
    3.10 3.10
    2.60 2.60
    3.60 3.60
    3.60 3.60
    3.10 3.10
    4.10 3.50
    3.50 3.50
    3.00 3.50
    4.00 3.50
    3.50 3.50
    3.00 3.50
    4.00 3.50
    3.50 3.50
    3.00 3.00
    4.00 3.40
    3.50 3.50
    3.00 3.00
    4.00 3.50
    1.70 1.70
    2.70 2.70
    2.20 2.20
    3.20 3.50
    3.50 3.50
    3.00 3.00
    4.00 3.40
    2.90 2.90
    2.40 2.40
    3.40 3.40
    3.50 3.50
    3.00 3.00
    4.00 3.70
    3.50 3.50
    3.00 3.50
    4.00 3.60
    2.20 2.00
    1.70 3.50
    2.70 2.70
    1.70 1.70
    1.20 1.20
    2.20 2.00
    3.50 3.50
    3.00 3.00
    4.00 3.50
    3.50 3.50
    3.00 3.00
    4.00 3.60
    3.50 3.50
    3.00 3.00
    4.00 3.50
    3.60 3.60
    3.10 3.10
    4.10 3.50
    3.60 3.60
    3.10 3.10
    4.10 3.50
    2.00 2.00
    1.50 3.10
    2.50 2.50
    3.50 3.50
    3.00 3.00
    4.00 3.60
    3.50 3.50
    3.00 3.00
    4.00 3.50
    3.50 3.50
    3.00 3.00
    4.00 3.50
    3.50 3.50
    3.00 3.50
    4.00 3.60
    2.60 2.60
    2.10 2.10
    3.10 3.10
    2.60 2.60
    3.60 3.60
    3.40 3.40
    2.90 2.90
    3.90 3.90
    };
    \addlegendentry{$\pm$0.5\,Ga record sweep}
  \end{axis}
\end{tikzpicture}
    \caption{\GVage{} open book, $\pm$0.5\,Ga record perturbation (slope \OBPertSlope{})}
    \label{fig:openbook-sweep}
  \end{subfigure}
  \caption{Stated age against reference or injected age on the held-out test split. (a) Closed book against centroid ground truth. Points cluster at the global mare prior, and the apparent residual skill (Table~\ref{tab:main-results}) is explained as memorized regional structure by the leave-one-region-out control (Table~\ref{tab:loo}). (b) Closed book against polygon-accurate unit ages. The stated values concentrate at the modal mare age, including for \GVchrono{}, whose targets state tile-specific chronology-consistent ages, while the same latent supports a positive-slope probe (Table~\ref{tab:age-bottleneck}). (c) Open book, the stated age against the injected dated record on the polygon set (Sec.~\ref{sec:res-openbook}). The stated values track the record along the identity, also where the record disagrees with the polygon ground truth (brighter points), so the residual error is inherited from retrieval. (d) Rewriting the injected record by $\pm$0.5\,Ga moves the stated age with slope \OBPertSlope{}, so the transfer is causal rather than a coincidence with the memorized regional prior of panel (a). The panels use different ground-truth conventions and tile sets and are not directly comparable to one another.}
  \label{fig:age-values}
\end{figure*}

Dating is where interpretation becomes quantitative, and here the balance between internalized knowledge and local evidence, posed in the introduction, tips toward the prior. Stated age values regress to the global mare prior under every target design, while probes recover the dating signal up to the resampler output, so the failure lies in the language-side extraction. Against centroid ground truth, the served model marginally surpasses the constant baseline (MAE \GVArMae{} vs.\ \GVConstMae{}\,Ga, Spearman $\rho=\GVSpearman$, Fig.~\ref{fig:age-values}a), but the difference of $-0.005$\,Ga falls well within the cluster-bootstrap interval of $[-0.022, +0.014]$\,Ga. A leave-one-region-out control eliminates the apparent skill. With all published ages for eastern Oceanus Procellarum omitted from training, the model performs nearly identically to the fully trained version on that region (age MAE 0.807 vs.\ 0.844\,Ga at comparable coverage), with both defaulting to approximately 3.45\,Ga for the genuinely younger basalts there (Table~\ref{tab:loo}). The nominal centroid-GT skill hence reflects a memorized regional age field rather than age values inferred from imagery.

\begin{table}[htbp]
  \centering
  \caption{Leave-one-region-out age recovery (eastern Oceanus Procellarum, closed-book, training-format prompts without retrieval text). The excised model never saw the zone's published ages in its corpus or targets, yet scores essentially the same as the fully trained model, and both default to $\sim$3.45\,Ga on genuinely younger basalts. The stated values track the global mare prior, not the imagery.}
  \label{tab:loo}
  \begin{tabular}{lccc}
    \toprule
    Model & Zone age coverage & Zone age MAE (Ga) & Epoch adjacent \\
    \midrule
    Trained with zone ages   & 0.800 & 0.844 & 1.000 \\
    Zone ages excised (LOO)  & 0.725 & 0.807 & 1.000 \\
    \bottomrule
  \end{tabular}
\end{table}

\GVchrono{}, whose targets are built on the crater chronology (Sec.~\ref{sec:method:targets}), is the strongest target-side design against this failure. Behavioral metrics improved, with age coverage rising from \PAKeeperCov{} to \PACraterCov{} on the polygon-GT set, yet the values themselves declined (Table~\ref{tab:age-poly}, Fig.~\ref{fig:age-values}b). Against polygon-accurate unit ages, \GVchrono{} yields an MAE of \PACraterMae{}\,Ga compared to \PAConstMae{}\,Ga for the constant prior, with $\rho=\PACraterRho{}$, and its reported crater densities show no correlation with the measured values. By contrast, \GVage{} shows a positive rank correlation (\PAKeeperRho{}), a near-zero slope (\PAKeeperSlope{}), and a pronounced bias toward older ages, indicative of a smooth, memorized age field. Neither generation channel outperforms the constant prior. Having exhausted the target-side interventions (emission gating, quantization granularity, token upweighting, and chronology-aligned supervision), we conclude that target design does not enable numeric age recovery in the generation channel.

\begin{table}[htbp]
  \centering
  \caption{Closed-book numeric age against polygon-accurate unit ages (point-in-polygon on the full mapped geometry, $n=93$ dated held-out tiles, identical seeded set for both models). Neither generation channel beats the constant prior on MAE. The \GVage{} targets teach a smooth regional age field (positive rank correlation, strong old bias), while the \GVchrono{} targets remove that crutch without transferring the counting signal. The frozen-latent probe that diagnoses this failure is reported in Table~\ref{tab:age-bottleneck} on its own evaluation set.}
  \label{tab:age-poly}
  \begin{tabular}{lccccc}
    \toprule
    & Coverage $\uparrow$ & MAE (Ga) $\downarrow$ & Spearman $\rho$ & Slope & Bias (Ga) \\
    \midrule
    \GVage{} (served) & 0.645 & 0.413 & 0.387 & 0.044 & 0.327 \\
    \GVchrono{} & 0.731 & 0.473 & -0.210 & -0.054 & 0.171 \\
    Constant prior (3.5\,Ga) & -- & 0.407 & 0 & 0 & -- \\
    \bottomrule
  \end{tabular}
\end{table}

\label{sec:res-gap}%
This failure is not attributable to missing visual information. A ridge probe on the frozen encoder latent, under a leave-unit-out evaluation (Appendix~\ref{app:probes}), recovers age with an MAE of \PAProbeMae{}\,Ga against \PAProbeConstMae{}\,Ga for the constant prior, with $\rho=\PAProbeRho{}$ and a slope of \PAProbeSlope{} (Table~\ref{tab:age-bottleneck}). Probes for log crater density perform better still, and shuffle controls collapse to the constant prior. The ceiling at this tile scale is itself modest. The true crater counts in place of vision reach an MAE of \PAProbeCeilMae{}\,Ga ($\rho=0.47$), a limit set by small-number crater statistics in 51$\times$67\,km windows, and the latent recovers just under a quarter of this margin (\PAProbeCeilFrac{}). A randomly initialized encoder over the same trained tokenizers probes nearly identically ($\rho=\PAProbeRandRho{}$), so the signal is carried by the tokenized input representation, which suffices for the diagnosis.

\begin{table}
  \centering
  \caption{Where the numeric dating signal is lost. The identical leave-unit-out probe battery
    ($20{,}956$ mare tiles from 290 dated units, identical folds) is applied at successive taps
    of one frozen forward pass. No bridge training and no language-model forward is involved.
    Row~1 probes the encoder latent, row~2 the mean-pooled resampler output that the in-loop age
    head reads, row~3 the same output without pooling. Each row reports the better of ridge and
    MLP by held-out MAE and carries a row-shuffle control that collapses toward the constant.
    Every probe faces the constant prior of this set (\PAProbeResConstMae{}\,Ga). The bottleneck
    passes the dating signal. This localizes the failure and claims no numeric age recovery by
    any generation model.}
  \label{tab:age-bottleneck}
  \begin{tabular}{lcccc}
    \toprule
    Probe input & MAE (Ga) $\downarrow$ & Spearman $\rho$ & Slope & Shuffled $\rho$ \\
    \midrule
    MAE encoder latent (768-d), condition A & 0.429 & 0.342 & 0.143 & -0.110 \\
    \quad $\to$ resampler, mean-pooled (3584-d), condition B & 0.428 & 0.306 & 0.126 & -0.121 \\
    \quad $\to$ resampler, all 64 tokens, condition B2 & 0.427 & 0.312 & 0.125 & -0.050 \\
    \midrule
    Constant prior (per fold) & 0.449 & 0 & 0 & -- \\
    \bottomrule
  \end{tabular}
\end{table}

\begin{figure}[htbp]
  \centering
  \begin{tikzpicture}
\begin{axis}[
    width=0.95\linewidth, height=0.42\linewidth,
    xlabel={training step}, ylabel={age MAE (Ga) $\downarrow$},
    label style={font=\small}, tick label style={font=\footnotesize},
    ymin=0, xmin=0, legend pos=north east, legend cell align=left,
    legend style={font=\footnotesize, draw=none, fill=none},
    grid=both, grid style={black!10},
]
    \addplot[thick, mark=*, mark size=1.2pt, color1!70!black] coordinates { (2000,0.535) (4000,0.488) (6000,0.401) (8000,0.558) (10000,0.413) (12000,0.373) (14000,0.368) (16000,0.370) };
    \addlegendentry{regression head (held-out)}
    \addplot[thick, mark=square*, mark size=1.2pt, color6!85!black] coordinates { (2000,0.529) (4000,0.512) (6000,0.438) (8000,0.426) (10000,0.402) (12000,0.387) (14000,0.371) (16000,0.374) };
    \addlegendentry{shuffled-vision control}
    \addplot[thick, dashed, color7!70!black] coordinates { (2000,0.260) (4000,0.260) (6000,0.260) (8000,0.260) (10000,0.260) (12000,0.260) (14000,0.260) (16000,0.260) };
    \addlegendentry{constant-prior baseline}
\end{axis}
\end{tikzpicture}
  \caption{Held-out age MAE of the in-loop supervised age head, a regression head trained jointly on the mean-pooled resampler output (mare-gated Huber loss), evaluated per training snapshot ($n=40$). The constant-prior baseline is recomputed on this evaluation's own set of $n=40$ dated snapshot tiles (0.26\,Ga MAE) and is therefore lower than the constants of Table~\ref{tab:main-results} and Table~\ref{tab:age-poly}, which live on different tile sets (Sec.~\ref{sec:eval:vlm}). The head never beats this baseline, and the shuffled-vision control tracks it closely. Together with the frozen-resampler probe (Table~\ref{tab:age-bottleneck}), the localization is that the representation the head reads provably carries the signal, and the joint training fails to extract it (Sec.~\ref{sec:res-gap}).}
  \label{fig:age-head}
\end{figure}
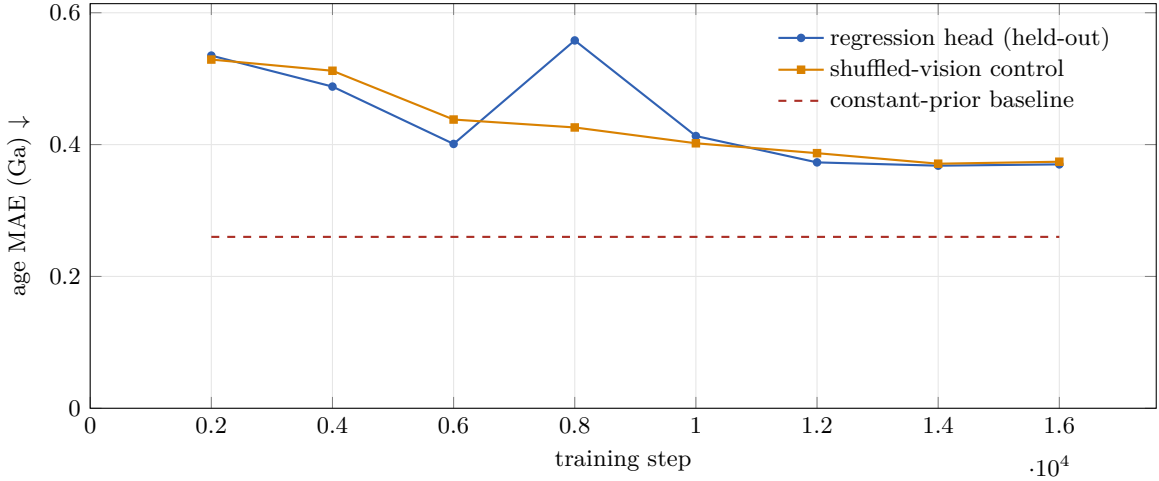

The probe on the frozen resampler output of the served bridge then localizes the loss. Under the same protocol, it recovers the signal at the level of the encoder probe (MAE \PAProbeResPooledMae{}\,Ga, $\rho=\PAProbeResPooledRho{}$, slope \PAProbeResPooledSlope{}), and probing all \PAProbeResK{} unpooled tokens excludes the pooling, so the bottleneck transmits the signal. An age head trained jointly on the same pooled vector, in contrast, never outperforms the constant prior, although shuffling its visual input degrades it substantially, so it reads vision but learns a regional-appearance proxy that fails across the geographic block split (Fig.~\ref{fig:age-head}). The unit-naming head on the same vector succeeds, so categorical identity is preserved. The same latent that supports a probe slope of \PAProbeSlope{} yields a slope of $-0.054$ through the generation channel, the probe-generation gap. The failure thus lies in extraction through the language pathway during joint training, rather than in representation itself.

\subsection{Retrieval supplies the chronology that vision cannot}
\label{sec:res-openbook}
A geologist does not date a 51$\times$67\,km scene from its crater counts alone (Sec.~\ref{sec:res-gap}) but consults the published chronology. The deployed configuration therefore enables retrieval, the open-book half of the regime comparison posed in the introduction. The stated age follows the injected record causally (perturbation slope \OBPertSlope{}), visually gated emission persists, and deployed age accuracy is bounded by the retrieval oracle rather than by the model. The served bridge was trained without retrieval text in its prompt, so the open-book context is a distribution shift at inference. The behavioral profile nevertheless carries over with minor variations (Table~\ref{tab:openbook}, block A), and causal grounding is preserved, with a tile-swap flip of \OBShuffle{} against \GVShuffle{} closed book. Although the geometric retrieval gate can pass a highland tile the age of a nearby dated mare polygon, the highland false-age rate rises only from \pct{\GVHl} to \pct{\OBHl}. The mare gate is a visual behavior, not a text reflex.

What retrieval changes is the stated value. On mare tiles with an injected record and a stated age ($n=167$), the stated value matches the retrieved one within 0.05\,Ga on \pct{\OBTransferExact} of tiles (mean absolute difference \OBTransferMae{}\,Ga, Fig.~\ref{fig:age-values}c). The injected centroid record coincides with the centroid ground truth, so this agreement is circular and is reported strictly as transfer fidelity, not as age recovery. In the non-circular polygon-GT set (Table~\ref{tab:openbook}, block B), open book performs comparably to the constant prior overall (MAE \OBPolyMae{} versus \OBConstMae{}\,Ga), reflecting imperfect retrieval rather than model inference. Where the injected record matches the polygon age ($n=\OBStratRightN$), the stated age is nearly exact (MAE \OBStratRightMae{}\,Ga), while mismatches are transferred equally faithfully (MAE \OBStratWrongMae{}\,Ga). The retrieval oracle, the injected record scored against the polygon ground truth, reaches \OBOracleMae{}\,Ga, so deployed age accuracy is bounded by retrieval quality rather than by the model.

Two interventions show that the transfer is causal rather than a coincidence with the memorized regional field of Sec.~\ref{sec:res-agevalues}. Rewriting the injected age by $\pm$0.5\,Ga moves the stated age with slope \OBPertSlope{} ($r=\OBPertR$, Fig.~\ref{fig:age-values}d), with a cluster interval excluding zero. Swapping in another tile's full context moves the stated age to the foreign record with slope \OBSwapSlope{}, while highland silence survives the foreign mare context. A geo-suppressed control, the same context without the dated record, leaves mare age coverage essentially unchanged. The record supplies the numeric value, whereas the decision to state an age remains the visually gated behavior of Sec.~\ref{sec:res-emission}.

Fig.~\ref{fig:example-openbook} illustrates the correction on a young Oceanus Procellarum tile whose record is accurate. Closed book, the model states a prior-shaped 3.1\,Ga and misplaces the scene. Open book, with identical raster-derived facts, it states the published 2.1\,Ga and the correct region. The generation channel cannot infer the age value from vision but transfers it faithfully from the retrieved text, so the deployed system states correct published ages wherever a valid record covers the tile. The two regimes delineate what the machine geologist reads from the site and what it must cite from the literature.
\begin{table*}
  \centering
  \caption{Open-book (deployed, retrieval-on) evaluation of the served model. Block A (centroid set): behavior is unchanged from closed book while the stated age transfers the injected record. Agreement with the injected record is circular and is reported only as transfer fidelity. Block B (polygon-accurate GT, non-circular): open book does not beat the constant prior overall because centroid retrieval is imperfect, and conditioning on record correctness decomposes this into near-exact accuracy where retrieval is right and record-tracking where it is wrong, so deployed accuracy is bounded by the retrieval oracle rather than by the model. The perturbation slope shows that the transfer is causal.}
  \label{tab:openbook}
  \begin{tabular}{lcc}
    \toprule
    \multicolumn{3}{l}{\textit{Block A, centroid set ($n=\OBN$): behavior and transfer fidelity}} \\
    \midrule
    Metric & Closed book & Open book \\
    \midrule
    Stated-fact fidelity (overall) $\uparrow$   & \GVGf{} & \OBGf{} \\
    Names mapped unit $\uparrow$                 & \GVNaming{} & \OBNaming{} \\
    Mare age coverage $\uparrow$                 & \GVCov{} & \OBCov{} \\
    Highland false-positive rate $\downarrow$    & \GVHl{} & \OBHl{} \\
    Consecutive-repeat rate $\downarrow$         & \GVCrep{} & \OBCrep{} \\
    Distinct-1 $\uparrow$                        & \GVDOne{} & \OBDOne{} \\
    \midrule
    Transfer fidelity (stated vs.\ injected record) & \multicolumn{2}{c}{exact \OBTransferExact{}, MAE \OBTransferMae{}\,Ga ($n=167$)} \\
    \midrule
    \multicolumn{3}{l}{\textit{Block B, polygon set ($n=\OBPolyN$): non-circular numeric age accuracy}} \\
    \midrule
    Condition & Coverage & MAE (Ga) $\downarrow$ \\
    \midrule
    \GVage{} closed book                      & 0.645 & \OBPolyClosedMae{} \\
    \GVage{} open book                        & \OBPolyCov{} & \OBPolyMae{} \\
    \quad where retrieval correct ($n=\OBStratRightN$)  & 0.603 & \OBStratRightMae{} \\
    \quad where retrieval wrong ($n=\OBStratWrongN$)      & 0.750 & \OBStratWrongMae{} \\
    Retrieval oracle (record vs.\ polygon GT) & 1.000 & \OBOracleMae{} \\
    Constant prior (\OBConstPrior{}\,Ga)  & -- & \OBConstMae{} \\
    \midrule
    Perturbation slope (stated on injected, $\pm$0.5\,Ga) & \multicolumn{2}{c}{\OBPertSlope{} ($r=\OBPertR$)} \\
    \bottomrule
  \end{tabular}
\end{table*}

\begin{figure*}
  \centering
  \input{figures/results/fig_example_openbook.tex}
  \caption{Closed book against open book on a held-out Oceanus Procellarum tile (long=35\textdegree{}38'9.33"W, lat=15\textdegree{}5'3.92"N, shown for the reader, not the model). Both answers are generated under the trained coordinate-free interpretation instruction, and the open-book prompt additionally carries the retrieved context, whose dated record (2.1\,Ga for unit P43, here consistent with the polygon ground truth) is the only age information available to either mode. The raster-derived opener facts are identical on both sides, with the same two hits and the same two misses against the measured values, so the context changes only the retrieved content, not the visual reading. The tile was selected from the candidate scan for this correction, and the answers are abridged at the marked ellipses. Table~\ref{tab:openbook} quantifies the transfer over all tiles.}
  \label{fig:example-openbook}
\end{figure*}

\section{Discussion}
\label{sec:discussion}

The results support one claim and its precise negation. Tile-grounded geological prose is achievable with a frozen, self-supervised vision tower and a standard bridge, provided the supervision targets cannot be reproduced without reading the tile. Near-identical prose targets produced a model that ignored the tile, deterministic raster-derived openers produced one that reads it, and coarsening those openers traded grounding away again. The same targets that control where and how the model speaks about age, however, do not make its stated age values track the surface. We consider the target-side avenue exhausted (Sec.~\ref{sec:res-agevalues}). The dissociation has a plausible mechanism. Token-level cross-entropy rewards reproducing the distributional properties of the targets, and emission behavior is such a property, whereas the numeric value inside an age sentence must be regressed from a continuous visual statistic, and for values concentrated near the modal mare age the loss incurred by always stating the prior is small. We note that this account is consistent with our evidence but was not tested in isolation.

The probe battery constrains where the information is lost. The dating signal enters the pipeline and survives the resampler bottleneck unchanged. What remains is the joint cross-entropy optimization, under which the in-loop age head settled on a regional-appearance proxy that anti-transfers across the geographic block split, and the language-side extraction it feeds. Two ceilings qualify how much numeric dating to expect at this scale. Tile-level crater statistics on 51$\times$67\,km windows are small-number statistics against unit-level ground truth, and even the true crater counts \citep{robbins2019new}, fed into the same probe in place of vision, yield only modest skill. Unit-level aggregation improved every probe we ran (age MAE \PAProbeUnitMae{}\,Ga, $\rho=\PAProbeUnitRho{}$). A decoupled dating head trained outside the language-modeling objective, reading a representation that the probes show to suffice, is the design our evidence indicates.

The open-book evaluation adds the deployment-side complement. The same generation channel that cannot regress the age value from vision transfers it faithfully and causally once the retrieval context contains the published record. In deployment, the numeric values are therefore already supplied by the retrieval corpus wherever a correct record covers the tile, and the immediate lever is retrieval accuracy, since a polygon-accurate lookup would remove the wrong-record stratum that dominates the deployed error. The measurements hence answer where scientific knowledge should reside in a machine geologist. Site evidence enters through vision and is verifiable against the rasters, the published chronology enters through retrieval and is citable, and the model's weights carry behavior, style, and the mare gate, but should not be trusted as a store of numeric ages, which the leave-one-region-out control shows to be memorized regional structure.

Beyond the system itself, the evaluation discipline transfers to any vision-language system trained on distilled targets over visually homogeneous scenes. Checkpoint selection must trade target adherence against causal grounding explicitly, the tile-swap ablation rather than vision-zeroing alone exposes tile-unspecific models, and degeneration statistics belong on the selection axis. LLaVA-LE \citep{LunarLLaVA}, the closest prior system, asserts rather than measures its grounding under a judge-based protocol (Sec.~\ref{sec:related}).

Several limitations bound the claims. The leave-one-region-out control was not re-run for the crater-chronology variant, so we make no leakage-free claim for it. The polygon-accurate age evaluation rests on 93 dated held-out tiles. The distilled targets originate from captions written by a single teacher model, which is why fit diagnostics carry no evidential weight anywhere in this paper, and the stylistic register inherits from one author's published corpus. The open-book results claim retrieval integration, never visual age recovery. All results concern one body, one tower, and one bridge, and systematic expert assessment of the generated interpretations remains to be done.

\section{Conclusions}
\label{sec:conclusion}

For the goal of scaling the interpretive layer of the Moon, the division of labor follows directly. Free-text generation, disciplined by target design and audited by causal ablation, closed-book scoring, and constant-prior baselines, is ready to carry the descriptive and stratigraphic content of an interpretive layer at the resolution of the data. Numeric chronology should not be generated but estimated, by decoupled heads reading the same frozen representations and aggregating over mapped units, and then cited by the language model like any other retrieved fact. Establishing that division quantitatively, rather than assuming either half, is what this work contributes.



\section*{Author contributions}
\textbf{Tom Sander}: Conceptualization, Data curation, Formal analysis, Investigation, Methodology, Project administration, Resources, Software, Supervision, Validation, Visualization, Writing - Original Draft, Writing - Review \& Editing; \textbf{Kay Wohlfarth}: Conceptualization, Validation, Writing - Review \& Editing; \textbf{Christian Wöhler}: Funding acquisition, Supervision, Validation, Writing - Review \& Editing;

\section*{Conflict of Interest}
The authors declare no conflicts of interest relevant to this study.

\section*{Open Research}
All input rasters are publicly archived products, namely the LROC WAC global mosaic \citep{wagner2015newWAC}, the GLD100 digital elevation model \citep{scholten2012gld100}, the Clementine UVVIS color-ratio composite \citep{eliason1999clementine,hare2008clementine}, the USGS Unified Geologic Map of the Moon \citep{fortezzo2020unified}, the Robbins lunar crater database \citep{robbins2019new}, and the IAU Gazetteer of Planetary Nomenclature \citep{iau_usgs_2026}, each cited in Sec.~\ref{sec:data:sources} with its source. The dated mare units and unit descriptions derive from the published literature cited in Secs.~\ref{sec:method:stageB} and~\ref{sec:method:targets}.

The training, evaluation, and asset-generation code is available at \url{https://github.com/TechnicToms/SelenoVLM}. The trained model weights are available at \url{https://huggingface.co/TechnicToms/SelenoVLM}.

\bibliographystyle{unsrtnat}
\bibliography{bibliography/lunar-geologic-mapping,bibliography/ml-plantary-surfaces,bibliography/multimodal,bibliography/geospatial,bibliography/coupling-vision-to-LLMs,bibliography/remote-sensing-vlm,bibliography/retrieval,bibliography/dataset,bibliography/methods,bibliography/introduction}

\clearpage
\setcounter{section}{0}
\setcounter{figure}{0}
\setcounter{table}{0}
\setcounter{equation}{0}
\renewcommand{\thesection}{\Alph{section}}
\renewcommand{\thefigure}{\Alph{section}\arabic{figure}}
\renewcommand{\thetable}{\Alph{section}\arabic{table}}
\renewcommand{\theequation}{\Alph{section}\arabic{equation}}
\makeatletter
\@addtoreset{figure}{section}
\@addtoreset{table}{section}
\@addtoreset{equation}{section}
\makeatother
\renewcommand{\theHsection}{appendix.\Alph{section}}
\renewcommand{\theHfigure}{appendix.\Alph{section}.\arabic{figure}}
\renewcommand{\theHtable}{appendix.\Alph{section}.\arabic{table}}
\renewcommand{\theHequation}{appendix.\Alph{section}.\arabic{equation}}

\section{Data preprocessing details}
\label{app:data}
For the spatially disjoint block split, the global grid is subdivided into macro-blocks of \SI{4096}{pixel} per side, and each block is deterministically assigned to a partition using a coordinate-based hash to guarantee reproducibility. Only tiles fully contained within a single partition are retained, while those that cross partition boundaries are discarded, ensuring zero pixel overlap between partitions.

\begin{figure}[htbp]
    \centering
    \input{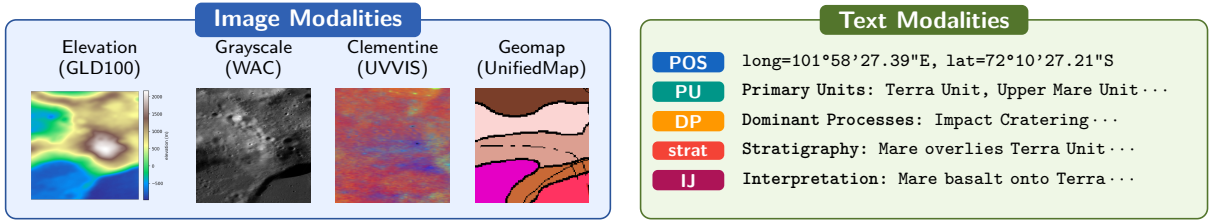}
    \caption{Overview of the input modalities. All raster products are co-registered at \SI{100}{\metre\per pixel}. The text modalities are derived from per-tile coordinates and from weakly supervised captions.}
    \label{fig:data:inputs}
\end{figure}

The geological map is supplied as an RGB raster with a categorical legend. Each pixel is mapped to the nearest legend color in RGB space and assigned the corresponding geological unit. This procedure produces a one-hot-encoded label volume spanning $50$ units (including background). Fig.~\ref{fig:data:inputs} renders the resulting input modalities.

\section{Tokenizer details}
\label{app:tokenizers}
Each raster tokenizer uses a Vision Transformer encoder with patch size $16$, so a $256\times 256$ single- or multi-channel tile corresponds to $256$ tokens. FSQ constrains and rounds each latent dimension to a fixed number of levels. The codebook is therefore the Cartesian product of the per-dimension level counts, which obviates the need for a learned codebook or an auxiliary commitment loss. The resulting code indices constitute the token vocabulary ($3^{8}=6561$ codes for WAC and elevation, $8\cdot8\cdot8\cdot6\cdot5=15{,}360$ for the Clementine composite, $4\cdot3^{6}=2916$ for the geological map). The associated decoder is a conditional diffusion model~\citep{ho2020denoising}. A U-Net \citep{ronneberger2015u} is optimized to denoise the target raster conditioned on the quantized codes, employing a noise-prediction objective that combines $\ell_1$ and $\ell_2$ losses with an LPIPS perceptual loss~\citep{zhang2018unreasonable}.

For the text modalities, every sequence is wrapped in \token{[SOS]}/\token{[EOS]} markers, padded with \token{[PAD]} or truncated to a fixed per-modality length ($64$ tokens for POS, $128$ for stratigraphy), with out-of-vocabulary items mapped to \token{[UNK]}. A reserved range of sentinel tokens (ids at or above a per-modality offset) marks the masked spans consumed by Stage~A. In the character-level POS tokenization, each digit and symbol (\texttt{=}, \texttt{\textdegree}, \texttt{'}, \texttt{"}, \texttt{.}, \texttt{,}, and a literal space) is its own token while contiguous letters are grouped:
\begin{center}\small
    \token{[SOS]}\token{long}\token{=}\token{1}\token{3}\token{6}\token{\textdegree}%
    \token{5}\token{6}\token{'}\token{3}\token{8}\token{.}\token{5}\token{7}\token{"}%
    \token{W}\token{,}\token{\textvisiblespace}\token{lat}\token{=}\token{1}\token{1}%
    \token{\textdegree}\token{2}\token{9}\token{'}\token{1}\token{0}\token{.}\token{3}%
    \token{4}\token{"}\token{N}\token{[EOS]}
\end{center}
In the word-level caption tokenization, a fixed keyword prefix encoding the section header is prepended, and the body is whitespace-split with punctuation detached as separate tokens. A stratigraphy field beginning \texttt{The Plains Unit overlies the Basin Lineated Unit, indicating\,\dots}\ yields:
\begin{center}\small
    \token{[SOS]}\token{**Stratigraphy}\token{(Relative}\token{Timing):**}\token{The}%
    \token{Plains}\token{Unit}\token{overlies}\token{the}\token{Basin}\token{Lineated}%
    \token{Unit}\token{,}\token{indicating}\ $\cdots$\ \token{[EOS]}
\end{center}

\section{Training configuration}
\label{app:training}
All training uses \texttt{bf16} mixed precision and the AdamW optimizer \citep{loshchilov2017decoupled} with cosine-annealed learning rates (minimum learning rate set to $5\%$ of the peak) and gradient clipping at $1.0$. Tiles are extracted at $512$~px with a stride of $128$~px and resized to the tokenizer input resolution. Hyperparameters for the two stages are listed in Tables~\ref{tab:impl-stageA} and~\ref{tab:impl-stageB}.

The Clementine stream was added to the trained eight-modality MAE via freeze-warmup head-add. During the first $2{,}000$ steps only the Clementine embeddings and output head are trained, after which all parameters are unfrozen. To decrease text-only repetition, optional penalties such as unlikelihood loss~\citep{welleck2019neural} and logit-space penalties (which discourage repeated outputs at inference) are implemented, although these are not used in the present configuration.

\begin{table}[h]
    \centering
    \begin{tabular}{@{}ll@{}}
        \toprule
        Hyperparameter & Value \\
        \midrule
        Modalities                 & WAC, elevation, Clementine, geomap, POS, and 4 caption fields (9 total) \\
        Transformer width $D$      & 768 \\
        Encoder / decoder depth    & 12 / 12 \\
        Attention heads            & 12 \\
        Feed-forward               & SwiGLU~\citep{shazeer2020glu}, no bias \\
        Drop-path rate             & 0.1 \\
        Embedding init.\ std.      & 0.15 \\
        Encoder / target token budget & 640 / 512 \\
        Modality-dropout prob.\    & 0.6 \\
        Optimizer                  & AdamW, $\beta=(0.9,0.99)$, weight decay $0.05$ \\
        Peak learning rate         & $1\times10^{-5}$ \\
        LR schedule                & cosine ($T_{\max}=2\times$ run length) \\
        Batch size                 & 34 \\
        Epochs                     & 180 \\
        Precision / devices        & \texttt{bf16}-mixed / 2 GPUs (DDP) \\
        \bottomrule
    \end{tabular}
    \caption{Stage~A (multimodal masked pre-training) configuration.}
    \label{tab:impl-stageA}
\end{table}

\begin{table}[h]
    \centering
    \begin{tabular}{@{}ll@{}}
        \toprule
        Hyperparameter & Value \\
        \midrule
        Language model             & Qwen2.5-7B-Instruct (frozen base) \\
        Resampler latents $K$      & 64 \\
        Resampler depth / heads    & 2 / 8 \\
        Gated cross-attention      & every 4th layer (7 blocks), 8 heads, $\tanh$ gate init.\ $0.1$ \\
        Cross-attention feed-forward & disabled \\
        LoRA rank / $\alpha$ / dropout & 16 / 32 / 0.05 \\
        LoRA target modules        & q, k, v, o, gate, up, down projections \\
        Optimizer                  & AdamW, $\beta=(0.9,0.95)$, weight decay $0$ \\
        Peak learning rate         & $1\times10^{-4}$ \\
        LR schedule                & cosine \\
        Batch size / grad.\ accum. & 4 / 4 (effective 16) \\
        Max sequence length        & 1024 \\
        Epochs                     & 5 \\
        Memory                     & gradient checkpointing, \texttt{bf16}-mixed, 1 GPU \\
        \bottomrule
    \end{tabular}
    \caption{Stage~B (SelenoVLM) configuration.}
    \label{tab:impl-stageB}
\end{table}

\section{Retrieval corpus and crater-chronology target details}
\label{app:corpus-chrono}
The retrieval corpus of Sec.~\ref{sec:method:stageB} comprises $13{,}007$ records, namely the geological-map unit descriptions, $291$ dated mare units with their mapped polygon geometries, partitioned passages from the source literature, and $9{,}085$ named features from the IAU gazetteer \citep{iau_usgs_2026}, all embedded~\citep{chen2024bge} and indexed for cosine similarity search~\citep{johnson2019billion}. A deterministic prose template constructed from the four caption fields serves as a fallback where no distilled target exists, and if no corpus is available, the retriever returns an empty context, allowing training to continue without retrieval augmentation.

The published model ages stated by the crater-chronology targets of Sec.~\ref{sec:method:targets} derive from the lunar chronology function of \citet{neukum2001cratering},
\begin{equation}
  N(1) = 5.44 \times 10^{-14} \bigl[ \exp(6.93\,T) - 1 \bigr] + 8.38 \times 10^{-4}\, T,
  \label{eq:chronology}
\end{equation}
which relates the cumulative density $N(1)$ of craters with diameters of at least \SI{1}{\kilo\metre}, per \si{\kilo\metre\squared} of counting area, to the absolute model age $T$ in Ga, calibrated against radiometrically dated Apollo and Luna samples. The stated tile density differs from $N(1)$ only in the areal normalization. We deliberately do not invert Eq.~\eqref{eq:chronology} on the tile densities. The published measurement protocol counts craters on geologically homogeneous areas and excludes secondary craters \citep{hiesinger2011ages}, whereas the tile densities are raw catalog counts on fixed 51$\times$67\,km windows, so a direct inversion would assign the off-polygon tiles ages systematically inconsistent with the published ages stated on the in-polygon tiles. The isotonic fit of Sec.~\ref{sec:method:targets} preserves the monotone increase of crater density with age that Eq.~\eqref{eq:chronology} implies, while calibrating the raw densities to the published record.

\section{Probe battery}
\label{app:probes}
The probe battery of Sec.~\ref{sec:res-gap} quantifies how much of the dating signal each stage of the pipeline retains. It is evaluated on its own set of $20{,}956$ mare tiles from $290$ dated units, where each tile carries the published age of its mapped unit as reference. On this set, the protection against memorized regional structure is provided by the cross-validation scheme rather than by the geographic block split. Writing $u(t)$ for the dated unit of tile $t$ and $z_t$ for the frozen MAE encoder latent of the tile, mean-pooled over the encoder token sequence to a fixed-length feature vector, each probe is a regression fitted and evaluated leave-unit-out,
\begin{equation}
  \hat{a}_t = g_{-u(t)}(z_t),
  \label{eq:loo-probe}
\end{equation}
where $g_{-u}$ denotes the regressor fitted on the tiles of all dated units except $u$. Every prediction is therefore out-of-unit, and no dated unit contributes to both fitting and evaluation. The probe family comprises ridge regressors and small multilayer perceptrons. The ridge regularization is selected within each training fold over a four-value grid ($\lambda \in \{10, 10^{2}, 10^{3}, 10^{4}\}$) on a held-out portion of the fold and refit with the winning value, with features standardized on training-fold statistics. The held-out portion holds out whole training groups (dated units for the age probe, spatial blocks for the density probe), so the selection of $\lambda$ is itself group-aware. The perceptron runs in one fixed configuration without any search. MAE, Spearman rank correlation, and regression slope, as defined in Sec.~\ref{sec:eval:vlm}, are computed over the pooled out-of-unit predictions.

Two probe targets are used. The age probe regresses the unit's published model age and reaches an MAE of \PAProbeMae{}\,Ga against the constant prior of \PAProbeConstMae{}\,Ga recomputed on the identical set, with $\rho = \PAProbeRho{}$ and slope \PAProbeSlope{} for the ridge, the better of the two families by held-out MAE (Table~\ref{tab:age-bottleneck}). The density probe regresses the logarithm of the tile's Robbins crater density and reaches a rank correlation of \PAProbeDensRhoTile{} per tile and \PAProbeDensRhoUnit{} after unit-level aggregation. Aggregation, that is, averaging the out-of-unit predictions over all tiles of a unit before scoring, improves every probe, and the aggregated age probe reaches an MAE of \PAProbeUnitMae{}\,Ga with $\rho = \PAProbeUnitRho{}$. Two controls guard the battery. The constant prior is recomputed on each probe's own evaluation set, and a shuffle control refits Eq.~\eqref{eq:loo-probe} after permuting the pairing between latents and reference ages, which collapses the probes to the constant prior and rules out leakage through the fitting protocol. The ceiling estimate of Sec.~\ref{sec:res-gap} replaces $z_t$ with the tile's true Robbins crater density under the identical leave-unit-out scheme, yielding an MAE of \PAProbeCeilMae{}\,Ga and $\rho = 0.47$.

The frozen-resampler probe of Table~\ref{tab:age-bottleneck} applies the identical battery, with byte-identical folds and the same controls, to two further taps of the same frozen forward pass of the served bridge, recorded by checkpoint hash. Condition B replaces $z_t$ with the mean-pooled resampler output (dimension \PAProbeResDim{}), the exact vector the in-loop age head reads. Condition B2 probes the unpooled \PAProbeResK{} resampler tokens under three views, the per-token statistics, a fixed-seed random projection of the flattened tokens, and a learned single-query attention pooling whose uniform special case reproduces mean pooling. The preservation criterion was fixed relative to condition A before any condition-B number existed. A family counts as preserved if its best model by held-out MAE beats the per-fold constant and reaches at least $0.7$ times the rank correlation and $0.5$ times the slope of condition A, with the shuffle control collapsed. Both conditions pass. No bridge parameter is updated at any point.

The in-loop age head of Fig.~\ref{fig:age-head} is not part of this battery. It is a LayerNorm followed by a small multilayer perceptron on the mean-pooled resampler output, trained jointly with the bridge at loss weight $0.5$ under a Huber loss, gated to tiles with a mare fraction of at least one half and a dated reference within $8^{\circ}$ (Sec.~\ref{sec:method:stageB}). Its per-snapshot evaluation uses $n=40$ dated snapshot tiles with the constant prior recomputed on that set.

\section{Vision-tower predictability}
\label{app:mae-fig}
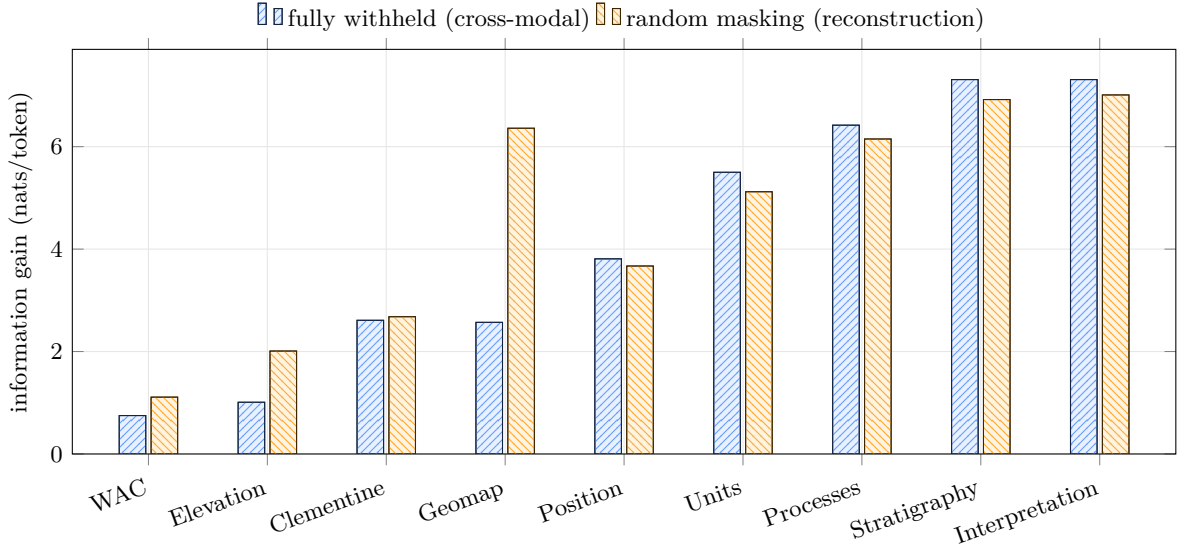
\begin{figure}[h]
  \centering
  \begin{tikzpicture}
\begin{axis}[
    width=0.98\linewidth, height=0.42\linewidth,
    ybar, bar width=10pt, ymin=0, ymax=7.9,
    enlarge x limits=0.08,
    symbolic x coords={WAC,Elevation,Clementine,Geomap,Position,Units,Processes,Stratigraphy,Interpretation}, xtick=data,
    x tick label style={font=\footnotesize, rotate=20, anchor=north east},
    ylabel={information gain (nats/token)}, label style={font=\small},
    y tick label style={font=\footnotesize},
    legend style={font=\footnotesize, at={(0.5,1.02)},
    anchor=south, legend columns=2, draw=none, fill=none},
    grid=major, grid style={black!10},
]
    \addplot[
      draw=color1!25!black,
      preaction={fill=color1!35!white, fill opacity=0.33},
      pattern=north east lines,
      pattern color=color1!80
    ] coordinates { (WAC,0.75) (Elevation,1.01) (Clementine,2.61) (Geomap,2.57) (Position,3.81) (Units,5.50) (Processes,6.42) (Stratigraphy,7.31) (Interpretation,7.31) };
    \addlegendentry{fully withheld (cross-modal)}
    \addplot[
      draw=color6!25!black,
      preaction={fill=color6!35!white, fill opacity=0.33},
      pattern=north west lines,
      pattern color=color6!80
    ] coordinates { (WAC,1.11) (Elevation,2.01) (Clementine,2.68) (Geomap,6.36) (Position,3.67) (Units,5.12) (Processes,6.15) (Stratigraphy,6.92) (Interpretation,7.01) };
    \addlegendentry{random masking (reconstruction)}
\end{axis}
\end{tikzpicture}
  \caption{Per-modality predictability in one comparable unit, computed as the uniform-vocabulary NLL floor minus the measured token NLL (nats per token), with the target modality fully withheld (cross-modal) or partially masked as in training (reconstruction). Structured modalities (captions, geologic map) are largely recoverable from the co-registered context. Raster token identity is not, although decoded rasters remain perceptually faithful (Table~\ref{tab:mae-tower}).}
  \label{fig:mae-tower}
\end{figure}

Fig.~\ref{fig:mae-tower} complements Table~\ref{tab:mae-tower} with the per-modality predictability in one comparable unit. The caption sections gain 5.5 to 7.3 nats/token over the uniform floor in the cross-modal regime, yet free-running BLEU-4 stays at or below 0.17 with no exact matches, and the shortfall is not repetition (distinct-1 $\geq 0.96$). The tower reproduces the section format and the dominant regional content but not the tile-specific wording, which is the circularity that the grounded targets of Sec.~\ref{sec:method:targets} break.

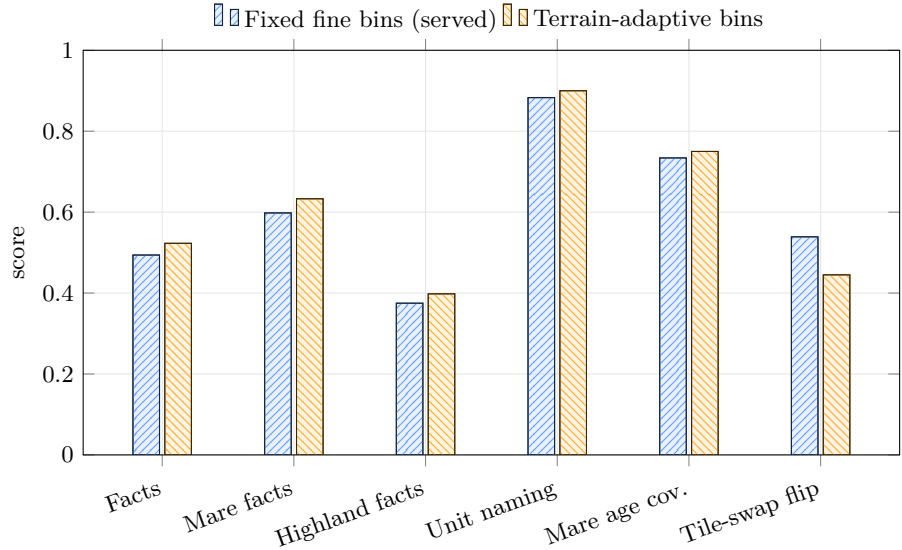
\begin{figure}[h]
  \centering
  \begin{tikzpicture}
\begin{axis}[
    width=0.75\linewidth, height=0.42\linewidth,
    ybar, bar width=10pt, ymin=0, ymax=1.0,
    enlarge x limits=0.12,
    symbolic x coords={Facts,Mare facts,Highland facts,Unit naming,Mare age cov.,Tile-swap flip}, xtick=data,
    x tick label style={font=\footnotesize, rotate=20, anchor=north east},
    ylabel={score}, label style={font=\small},
    y tick label style={font=\footnotesize},
    legend style={font=\footnotesize, at={(0.5,1.02)},
    anchor=south, legend columns=2, draw=none, fill=none},
    grid=major, grid style={black!10},
]
    \addplot[
      draw=color1!25!black,
      preaction={fill=color1!35!white, fill opacity=0.33},
      pattern=north east lines,
      pattern color=color1!80
    ] coordinates { (Facts,0.494) (Mare facts,0.598) (Highland facts,0.375) (Unit naming,0.883) (Mare age cov.,0.734) (Tile-swap flip,0.539) };
    \addlegendentry{Fixed fine bins (served)}
    \addplot[
      draw=color6!25!black,
      preaction={fill=color6!35!white, fill opacity=0.33},
      pattern=north west lines,
      pattern color=color6!80
    ] coordinates { (Facts,0.523) (Mare facts,0.633) (Highland facts,0.398) (Unit naming,0.900) (Mare age cov.,0.750) (Tile-swap flip,0.445) };
    \addlegendentry{Terrain-adaptive bins}
\end{axis}
\end{tikzpicture}
  \caption{Terrain-adaptive bins vs.\ the served fixed fine bins on the identical seeded held-out tiles (Table~\ref{tab:bins-ablation}). Adaptive bins win every content metric, highland fact fidelity most of all, but lose causal grounding (paired tile-swap $\Delta=-0.093$, $t=-3.90$), the axis on which the serving decision is gated.}
  \label{fig:bins-ablation}
\end{figure}

\section{Closed-book chat example}
\label{app:example}
Fig.~\ref{fig:example} shows the free-form question mode on a single held-out tile. It complements the instruction-format closed-book against open-book comparison of Fig.~\ref{fig:example-openbook}, which is the deployed configuration the quantitative evaluation scores. The model receives the four co-registered rasters, no coordinates and no retrieval context, and answers a free-form age question. All four opener facts it states fall within one scoring tolerance of the raster-measured values, the dominant mapped unit is named, and the interpretation argues from superposition relations between the mapped units rather than from a template. The tile was selected as a best-case age match, and the caption flags the two ungrounded statements in the answer.

\begin{figure*}[h]
  \centering
  \input{figures/results/fig_example.tex}
  \caption{Closed-book chat on a held-out tile (long=22\textdegree{}58'20.68"W, lat=55\textdegree{}39'24.92"N, shown for the reader, not the model). \GVage{} reads the four co-registered rasters, no coordinates and no retrieval context. The chips compare stated against raster-measured values and verify the four opener facts, the named unit, and the stated age, and nothing else. Two statements in the answer are not grounded. The place name \str{Schroter U} cannot come from the inputs, since the feature lies roughly $53^{\circ}$ of latitude away, and the phrase \str{regional crater statistics} asserts a measurement the model does not perform. Both are learned prose patterns of the target distribution. The WAC panel is rendered at natural reflectance, so the dark mare surface appears dark, consistent with the stated albedo, while the elevation and Clementine panels are contrast-stretched for legibility. The tile was selected as a best-case age match. Table~\ref{tab:age-poly} quantifies that stated age values in general track the mare prior rather than the tile.}
  \label{fig:example}
\end{figure*}

\end{document}